\documentclass[letterpaper]{article} 
\usepackage[]{aaai2026}  
\usepackage{times}  
\usepackage{helvet}  
\usepackage{courier}  
\usepackage[hyphens]{url}  
\usepackage{graphicx} 
\usepackage{natbib}  
\usepackage{caption} 
\usepackage{algorithm}
\usepackage{algorithmic}
\usepackage{booktabs}
\usepackage{multirow}
\usepackage{xcolor}
\usepackage{amsmath}
\usepackage{amsfonts}
\usepackage{dsfont} 
\usepackage{tcolorbox}
\usepackage{newfloat}
\usepackage{listings}
\DeclareCaptionStyle{ruled}{labelfont=normalfont,labelsep=colon,strut=off} 
\floatstyle{ruled}
\newfloat{listing}{tb}{lst}{}
\floatname{listing}{Listing}
\title{Logic Unseen: Revealing the Logical Blindspots of Vision-Language Models} 

\author{
Yuchen Zhou, Jiayu Tang, Shuo Yang, Xiaoyan Xiao, Yuqin Dai, Wenhao Yang, \\ Chao Gou, Xiaobo Xia, Tat-Seng Chua
}
\affiliations{
    Sun Yat-sen University, National University of Singapore
}

\begin{document}

\maketitle

\begin{abstract}
Vision-Language Models (VLMs), exemplified by CLIP, have emerged as foundational for multimodal intelligence. However, their capacity for logical understanding remains significantly underexplored, resulting in critical ``logical blindspots'' that limit their reliability in practical applications. To systematically diagnose this, we introduce LogicBench, a comprehensive benchmark with over 50,000 vision-language pairs across 9 logical categories and 4 diverse scenarios: images, videos, anomaly detection, and medical diagnostics.
Our evaluation reveals that existing VLMs, even the state-of-the-art ones, fall at over 40 accuracy points below human performance, particularly in challenging tasks like Causality and Conditionality, highlighting their reliance on surface semantics over critical logical structures. 
To bridge this gap, we propose LogicCLIP, a novel training framework designed to boost VLMs' logical sensitivity through advancements in both data generation and optimization objectives. 
LogicCLIP utilizes logic-aware data generation and a contrastive learning strategy that combines coarse-grained alignment, a fine-grained multiple-choice objective, and a novel logical structure-aware objective.
Extensive experiments demonstrate LogicCLIP's substantial improvements in logical comprehension across all LogicBench domains, significantly outperforming baselines. Moreover, LogicCLIP retains, and often surpasses, competitive performance on general vision-language benchmarks, demonstrating that the enhanced logical understanding does not come at the expense of general alignment. 
We believe that LogicBench and LogicCLIP will be important resources for advancing VLM logical capabilities. The data, code, and models will be publicly released.
\end{abstract}

%

\section{I. Introduction}

\textit{\textbf{``Logic is the anatomy of thought.''} — John Locke} 

\noindent Logic, as the intrinsic structure of human reasoning, permeates every facet of how we perceive the world, communicate effectively, and solve complex problems~\cite{kowalski2011computational}.
In natural language, logical constructs such as conjunction, disjunction, negation, and causality play a crucial role in shaping meaning beyond superficial semantics.
For example, consider the medical instructions \textit{``Do not mix drug A \textbf{and} drug B''} versus \textit{``Do not mix drug A \textbf{or} drug B''}. A subtle change in the logical connective system radically alters the meaning, potentially affecting realistic vital decisions.
As multimodal AI systems are increasingly deployed in critical applications \cite{liu2025lamra,li2022clip,lu2024visual,zhou2024few, luo2025gui,guo2024drivinggen}, such logical understanding becomes essential. Vision-Language Models (VLMs) must not only perceive images and interpret texts but also understand the complex logical structures that connect them, thereby preventing potentially catastrophic real-world errors.

Although VLMs, exemplified by CLIP \cite{radford2021learning,cherti2023reproducible}, excel in image-text alignment and have served as foundational encoders for various applications, such as open-world perception \cite{wang2023exploring,Sun_2024_CVPR,girdhar2023imagebind,zhou2024hierarchical} and Multimodal Large Language Models (MLLMs) \cite{li2023blip,chen2024internvl,liu2024improved}, their ability to comprehend and reason over logical structures remains significantly underexplored.
Figure~\ref{fig1} presents a preliminary analysis revealing these concerning \textbf{logical blindspots} in CLIP. 
As demonstrated, we perturb correctly matched image-text pairs by modifying object categories, attributes, and logical structures within the captions to construct erroneous descriptions. 
Our findings indicate that CLIP is highly sensitive to changes in objects or attributes, resulting in a significant drop in CLIP scores. However, it struggles to distinguish subtle logical shifts that fundamentally alter the entire sentence’s meaning, sometimes even assigning higher matching scores to these logically incorrect descriptions.
From left to right in Figure \ref{fig1}, CLIP fails to capture subtle variations in logical structures such as causality, inclusion, comparison, and conjunction. These deficiencies suggest that CLIP relies heavily on superficial semantic alignment, often missing deeper logical structures within natural language, which leads to logical blindspots.

\begin{figure*}[t]
\centering
	\includegraphics[trim=0 5 0 5, clip=true, width=1\textwidth]{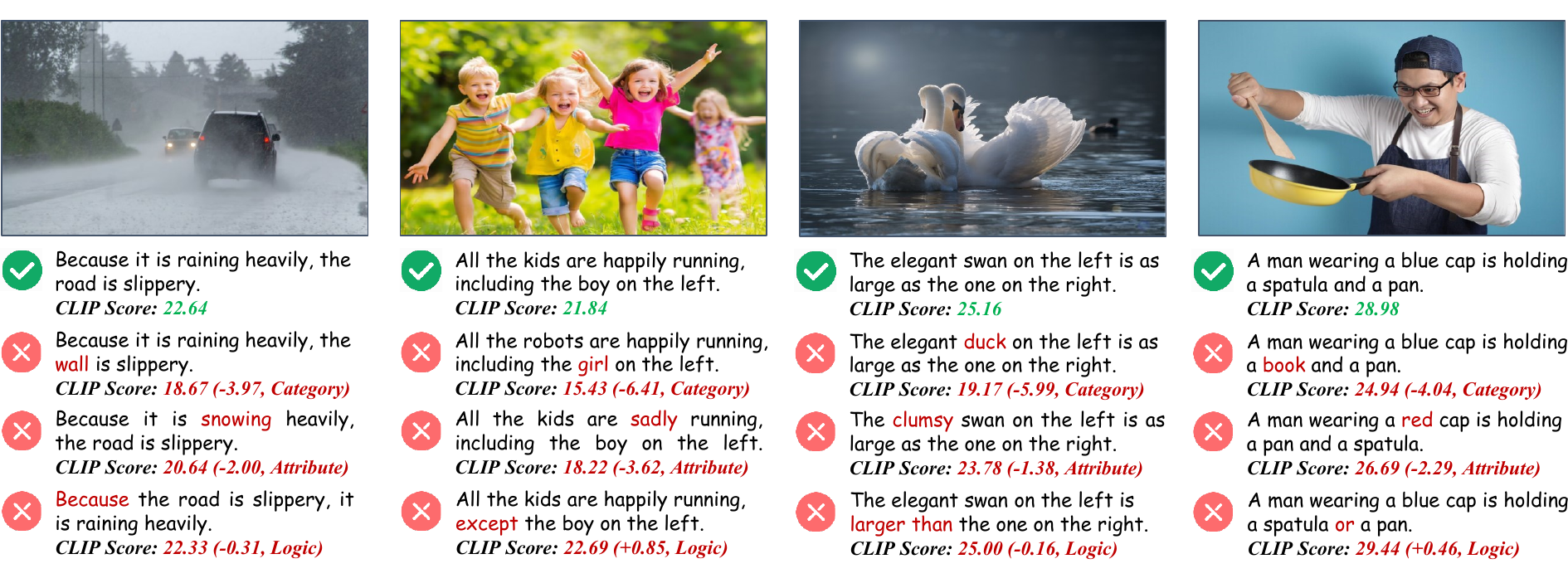}
	\caption{\textbf{Preliminary analysis of OpenAI CLIP-L/14 reveals logical blindspots.} While CLIP is highly sensitive to changes in objects or attributes, resulting in a significant drop in CLIP scores, it demonstrably struggles to distinguish subtle logical shifts that fundamentally alter the sentence's meaning, sometimes even assigning higher matching scores to descriptions containing logical errors. These examples include perturbations in causal, inclusion, comparison, and conjunctive logical relations.}  
 \label{fig1}
\end{figure*}

To bridge this gap, in this paper, we first introduce \textbf{LogicBench}, a novel and comprehensive benchmark specifically designed to systematically diagnose the logical understanding capabilities of VLMs. Specifically, LogicBench spans a wide range of domains, including everyday images, web videos, anomaly detection, and medical scenarios. It covers nine common logical categories, namely conjunction, disjunction, negation, contrast, comparison, condition, causality, temporality, and inclusion.
This broad coverage permits an in-depth evaluation of VLMs’ ability to comprehend logic across various contexts. Moreover, we propose two diagnostic tasks to probe these capabilities: (1) Logic-Aware Retrieval, which requires models to understand the logical structure of sentences and retrieve images from diverse data that match specific logical relations, simulating real-world tasks such as search engines, content moderation, and recommendation systems; and (2) Logic Multiple-Choice Questions (MCQ), which challenges models to perform fine-grained reasoning on subtitles with adversarial logical perturbations, selecting the most accurate description of an image from closely related options.

Furthermore, we propose \textbf{LogicCLIP}, a novel training framework that elevates the logical sensitivity of VLMs through logic-aware contrastive learning. Technically, we introduce a large-scale hard negative sample generation pipeline, where Large Language Models (LLMs) are employed to craft logically perturbed negative captions from original human-written descriptions meticulously. These captions serve as crucial supervisory signals during contrastive fine-tuning, enabling the model to better align visual content with logically coherent language. Moreover, our proposed logic-aware learning approach encourages the model not only to focus on surface-level semantics but also to explicitly attend to critical logical structures. Extensive experimental results demonstrate that while existing VLMs exhibit significant limitations in logical reasoning, our LogicCLIP not only achieves state-of-the-art performance on LogicBench but also maintains, and in some cases surpasses, the performance of native CLIP on general vision-language benchmarks. Before delving into details, we summarize our contributions as follows:


\begin{itemize}
    \item \textbf{(Benchmark)} We introduce LogicBench, a comprehensive benchmark specifically designed to systematically evaluate the logical understanding capabilities of current VLMs. LogicBench features over 50,000 logical vision-language pairs across 9 categories of logical relationships, encompassing 4 diverse application scenarios and 2 distinct evaluation tasks.
    \item \textbf{(Diagnosis)} We conduct the first systematic evaluation of VLMs' logical reasoning abilities, revealing significant ``logical blindspots" and inherent limitations in comprehending complex logical structures.
    \item \textbf{(Solution)} We present LogicCLIP, a novel and effective training framework engineered to boost VLMs' logical sensitivity. Extensive experiments validate that LogicCLIP achieves substantial improvements in logical comprehension across various domains while simultaneously preserving, and often surpassing, competitive performance on standard vision-language benchmarks.
\end{itemize}


\begin{figure*}[h]
\centering
	\includegraphics[trim=0 5 0 5, clip=true, width=1\textwidth]{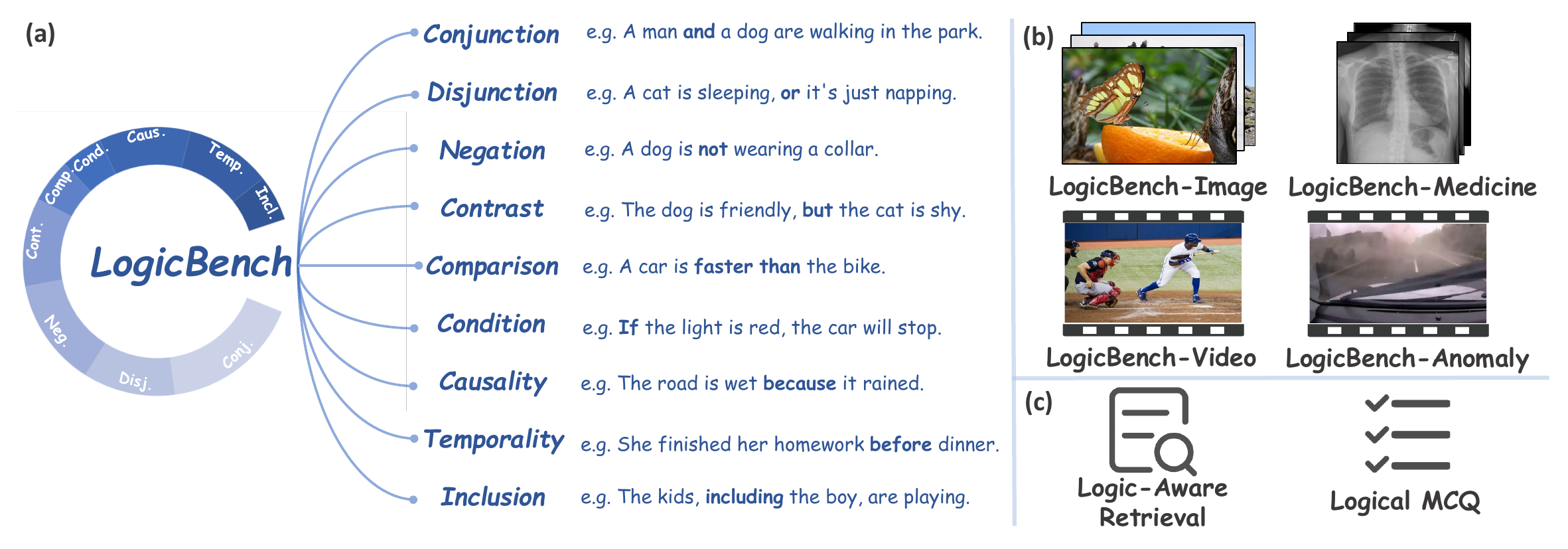}
	\caption{\textbf{The overview of our LogicBench.} A comprehensive benchmark for evaluating logical reasoning across various domains, including 9 common logical categories, 4 key application scenarios, and 2 challenging tasks.}
 \label{fig2}
\end{figure*}

\section{II. Related Work}

\textbf{Contrastive Language-Image Learning.}
CLIP \cite{radford2021learning} has demonstrated impressive zero-shot generalization capabilities by learning cross-modal alignment from massive web data, which has led to its widespread use in various downstream tasks such as recognition \cite{yu2025building,zhou2024learning}, detection \cite{esmaeilpour2022zero}, segmentation \cite{liu2025unveiling}, re-identification \cite{yu2024tf}, and diagnosis \cite{lu2024visual}. CLIP has also become the most commonly used vision encoder for recent MLLMs \cite{li2023blip,chen2024internvl,liu2024improved}, enabling LLMs to understand images. 
However, significant challenges persist. Recent works have highlighted CLIP's limitations in specific areas such as negation \cite{ko2025bringing,park2025know}, compositionality \cite{hsieh2023sugarcrepe}, and spatial reasoning \cite{wang2025spatialclip}, which hinder the reliability and performance of multimodal AI systems \cite{tong2024eyes,luo2025vcm,liu2024towards,zhou2023learning,liu2023learning,huang2024driver}.
Benchmarks such as CREPE \cite{ma2023crepe} and CC-Neg \cite{singh2025learning} have focused on the compositional understanding, relying on linguistic templates, but they lack diversity in real-world query phrasing. More recently, NegBench \cite{alhamoud2025vision} utilized LLaMA 3.1 \cite{grattafiori2024llama} to generate fluent negations, but the reliance on a single LLM introduces potential bias and hallucinations. 
As a comparison, we systematically diagnose the severe deficiencies of CLIP in understanding logical structures. Our proposed LogicBench includes 9 common logical categories, such as negation, causality, and contrast, and spans multiple visual domains such as natural images, videos, anomaly detection, and medical diagnostics. By incorporating multiple LLMs and human expert reviews to enhance data quality, LogicBench ensures diverse and high-quality logical reasoning tasks. This comprehensive strategy provides the community with a more robust and reliable benchmark for evaluating and improving the logical reasoning capabilities of VLMs.

\noindent\textbf{Synthetic Data for Model Training.}
With the growing prominence of LLMs, the use of synthetic data for training models has become a key area of exploration \cite{bauer2024comprehensive,yuan2024realfake,zhou2025towards,Li_2025_CVPR}. The generation of large-scale high-quality synthetic data has shown promise in enhancing model performance, especially when used to augment real data. Several works \cite{tian2024learning, liu2024can} have shown that models trained entirely on synthetic images and descriptions can achieve performance comparable to those trained on real-world data. 
In this paper, we leverage human-expert annotated logical captions as the foundation for generating challenging logic-violating hard negative samples through multiple LLMs. These synthetic data are then used to teach VLMs a higher-level skill—logical structure understanding.

\section{III. LogicBench}

To systematically diagnose the capability of VLMs to comprehend logical structures, we introduce LogicBench. Spanning diverse domains such as everyday images, web videos, anomaly detection, and medical scenarios, LogicBench covers 9 core logical categories and defines multi-granularity tasks to enable a comprehensive evaluation of VLMs’ logical reasoning across various real-world contexts. In this section, we detail the definition, statistics, and construction pipeline of LogicBench.

\subsection{Definition \& Statistics}
LogicBench is structured around three key dimensions: a defined set of logical categories, diverse application scenarios, and specifically designed logical-aware tasks for evaluation. 

\noindent \textbf{Logical Categories.}
LogicBench includes 9 fundamental logical categories, collectively forming 51,360 logical image-text pairs. Among these, the most frequent categories are conjunction (20.15\%), negation (15.63\%), and disjunction (13.36\%), while the less common categories include comparison (6.29\%) and condition (6.01\%), although they still provide a sufficient sample size. These categories enable a detailed analysis of VLMs in their ability to interpret complex logical relations beyond superficial matching. The specific categories and examples are illustrated in Figure \ref{fig2}(a).
\textit{See Supp. Mat. A for more details and statistics.}

\noindent \textbf{Application Scenarios.}
To comprehensively evaluate VLMs' logical understanding in real-world contexts, LogicBench incorporates data from four distinct scenarios: natural scene images, videos, anomaly detection, and medical diagnostics, as shown in Figure \ref{fig2}(b). 
Specifically,
\textbf{\textit{(1) Image}}: Sourced from the CC12M \cite{changpinyo2021conceptual} and MSCOCO \cite{lin2014microsoft} validation sets, these datasets feature highly diverse images from everyday environments.
\textbf{\textit{(2) Video}}: From the MSRVTT dataset \cite{xu2016msr}, containing popular video clips from a commercial video search engine, it is used to test the VLMs' ability to handle dynamic logical understanding.
\textbf{\textit{(3) Anomaly}}: Using the DADA-2000 dataset \cite{fang2019dada}, this scenario involves identifying traffic anomalies and accidents, requiring precise logical reasoning to identify unusual or hazardous events.
\textbf{\textit{(4) Medicine}}: Leveraging the Open-i dataset~\cite{demner2015preparing}, which includes radiology reports alongside medical images, this scenario demands nuanced logical interpretation for clinical relevance, such as understanding negation and causality in medical findings.

\noindent \textbf{Logical-aware Tasks.}
We design two diagnostic tasks with varying levels of granularity: Logic-Aware Retrieval and Logical Multiple Choice Questions (MCQ), as shown in Figure \ref{fig2}(c). Specifically,
\textbf{\textit{(1) Logic-Aware Retrieval}} requires the models to understand the logical structure of sentences and retrieve images from a diverse database that match specific logical relations. This simulates real-world applications such as search engines, content moderation, and recommendation systems. 
\textbf{\textit{(2) Logical MCQ}} challenges models to perform discriminative reasoning over subtitles containing adversarial logical perturbations. Models must select the most accurate description of a given image from a set of closely related options, requiring a precise understanding of subtle logical distinctions.
\textit{See Supp. Mat. A for visualizations.}

\subsection{Data Construction Pipeline}


Creating a large-scale, high-quality benchmark to diagnose VLMs' capabilities to understand logical structures presents two inherent challenges.
\textit{The first challenge is how to acquire a large-scale and logic-aware positive sample set.}
Directly using MLLMs to generate logical captions may introduce hallucinations and bias. These generated sentences are often crafted to fulfill the structural needs of logic, which can result in semantic misalignments with the corresponding visual content. While using human experts to annotate from scratch may serve as an alternative, the high cost associated with this approach severely limits the benchmark's scale. For example, the successful SpatialBench \cite{wang2025spatialclip}, despite considerable efforts, includes fewer than 300 MCQs.
To address this, we leverage existing human-annotated vision-language datasets and filter out captions containing logical structures to serve as our positive samples by utilizing spaCy and regular expression scripts for syntactic analysis. This approach enables us to process large datasets automatically, identifying samples with embedded logical relationships. Moreover, the selected samples are naturally annotated without the introduction of artificial complex logical structures, ensuring both the authenticity and practical relevance of the data.

\textit{The second challenge lies in generating a large-scale set of logic-aware negative samples based on the positive samples.} Previous works \cite{ma2023crepe,singh2025learning} often rely on template-based perturbations of captions, which could result in grammatically awkward sentences with low diversity and poor-quality negative samples. Other methods use a single LLM to generate negative samples \cite{alhamoud2025vision}, which addresses some of the issues but still introduces significant bias due to the lack of diversity.
Our solution involves using multiple widely validated LLMs (Qwen 2.5-max, DeepSeek-V3, Gemini-2.5-pro, GPT-4.1, and LLaMA 3.3 70B) to generate diverse and high-quality negative samples. This approach mitigates the bias introduced by relying on a single LLM. After generating both positive and negative samples, we engage human experts to review and select the most accurate and reliable samples, ensuring logical consistency and high quality. The dataset is then finalized, leading to LogicBench, which is a large-scale, high-quality benchmark that not only meets the scale requirements but also incorporates sufficient logical complexity, naturalness, and diversity to effectively diagnose current VLMs. 
Figure~\ref{fig3} illustrates the LogicBench construction pipeline, \textit{which is further detailed in Supp.Mat.A}.

\begin{figure}[t]
\centering
	\includegraphics[trim=15 5 15 5, clip=true, width=0.48\textwidth]{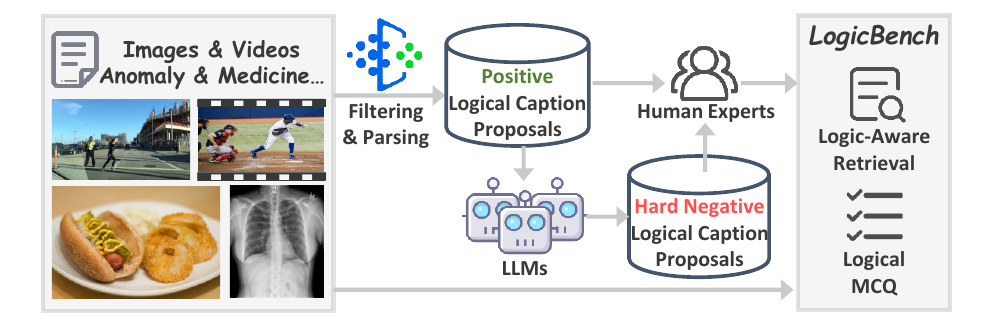}
	\caption{\textbf{LogicBench Construction Pipeline.}  We select vision-language pairs from human-annotated datasets, then parse and filter them for logical relations to form positive caption proposals. Multiple LLMs generate hard negative captions by perturbing the positive ones. Human experts review all proposals to create LogicBench.} 
 \label{fig3}
\end{figure}

\section{IV. LogicCLIP}

As noted, CLIP relies on superficial semantic alignment, matching image features with text based on simple co-occurrence patterns. However, it struggles to capture deeper logical structures, missing subtle semantic shifts in complex logical relationships. This leads to the phenomenon of ``logical blindspots''.
The root cause of these blindspots lies in both the \textit{training data} and \textit{optimization objectives}. 
First, the training data lacks sufficient explicit examples of complex logical relationships. Second, the optimization objective does not explicitly account for logical structures. 
The combination of factors prevents CLIP from learning these intricate logical relationships.
To address this, we propose LogicCLIP, which tackles ``logical blindspots'' challenges through both data augmentation and optimization. By integrating logic-aware training data generation and modifying the optimization objective to include logic-specific losses, LogicCLIP enhances VLMs' ability to capture complex logic, surpassing superficial semantic alignment.

\subsection{Logic-Aware Training Data Generation}

We expand the MSCOCO training set \cite{lin2014microsoft}, which includes high-quality human-annotated captions from daily image scenes, to generate our logic-aware training data. 
We aim to enhance the model's ability to adapt to diverse logical structures and comprehend logical relations in various scenarios. This is accomplished by utilizing naturally occurring logical sentence patterns from general daily contexts. The process involves two key steps:

\noindent \textbf{Positive Logical Sample Parsing.}
We begin by leveraging human-annotated captions from MSCOCO and applying spaCy and regular expression scripts for syntactic analysis. Let $\mathcal{I}$ be the set of images, $\mathcal{C}$ the set of captions, and $\mathcal{L} = \{L_1, \dots, L_k\}$ the set of predefined logical categories. For each image $I \in \mathcal{I}$ with associated captions $C_I \subset \mathcal{C}$, we filter and identify captions $c_{\text{pos}} \in C_I$ that contain specific logical structures. Each such caption $c_{\text{pos}}$ is then associated with a non-empty subset of logical categories $L \subseteq \mathcal{L}$. 
These $(I, c_{\text{pos}}, L)$ pairs form the positive samples, which the model is trained to recognize.
\textit{See Supp. Mat. B for more details.}

\noindent \textbf{Negative Logical Sample Generation.}
To create challenging negative samples, we generate a diverse set using multiple well-validated LLMs (Qwen 2.5-max, DeepSeek-V3, Gemini-2.5-pro, GPT-4.1, and LLaMA 3.3 70B).
For each positive sample $(I, c_{\text{pos}}, L)$, LLMs perturb $c$ to generate a set of three negative captions, denoted as $C_{\text{neg}} = \{c_{\text{neg},1}, c_{\text{neg},2}, c_{\text{neg},3}\}$. Each caption $c_{\text{neg},m} \in C_{\text{neg}}$ is semantically plausible but logically incorrect or misleading for image $I$ given the logical categories in $L$. This strategy ensures significant diversity in the negative sample set and pushes the model's capacity to differentiate subtle logical nuances.

The complete Logic-aware Training Data $\mathcal{D}$ consists of samples structured as $D = (I, L, c_{\text{pos}}, C_{\text{neg}})$, where each sample includes an image, its logical categories, a positive caption, and a set of three negative captions.
Notably, $\mathcal{D}$ is derived from general daily images and is not specifically curated for the four LogicBench scenarios. Nevertheless, as shown in Sec. V, LogicCLIP demonstrates remarkable generalization, performing robustly across logically critical domains such as video, anomaly detection, and medical diagnostics, even when fine-tuned solely on these daily images.

\subsection{Logic-Aware Contrastive Learning}

To equip VLMs with robust logical understanding capabilities, our optimization strategy integrates three distinct and complementary objectives: a coarse-grained standard CLIP contrastive objective, a fine-grained hard multiple-choice objective, and a novel logical structure-aware objective, which collectively enhance VLMs' understanding of logical relations within both visual and linguistic modalities.

\noindent \textbf{Standard CLIP Contrastive Objective.} 
The primary objective of this component is to improve the model's overall logic-aware visual-language alignment at a coarse-grained level. By ensuring that the model learns to align image content with logical text descriptions, this foundational objective sets the stage for more fine-grained logical understanding.
For a given batch \( \mathcal{B} = \{(I_i, C_i)\}_{i=1}^{N} \), where \( N \) is the number of image-caption pairs, we first extract their respective representations. An image encoder \( E_\text{I} \) produces image features \( \mathbf{v}_i = E_\text{I}(I_i) \), and a text encoder \( E_\text{T} \) generates text features \( \mathbf{t}_i = E_\text{T}(C_i) \). The cosine similarity matrix \( \mathbf{S} \in 
\mathbb{R}^{N \times N} \) is computed, where \( \mathbf{S}_{ij} = \cos(\mathbf{v}_i, \mathbf{t}_j) \). 
We apply a symmetric cross-entropy loss to this matrix, encouraging high similarity for correct image-caption pairs \( (\mathbf{v}_i, \mathbf{t}_i) \) and low similarity for incorrect pairs:
\[
L_{\text{CLIP}} = \mathcal{L}_{\text{CE}}(\mathbf{S}, \text{labels}),
\]
where labels indicate an identity matrix with ones on the diagonal indicating correct matches.

\noindent \textbf{Fine-Grained Multiple-Choice Objective.} 
This objective aims to improve the model’s ability to distinguish subtle logical differences, particularly when confronted with challenging negative samples. It directly leverages the image-positive sample-negative sample sets generated during data creation.
Given a batch \( \mathcal{B}_{\text{MC}} = \{(I_i, c_{\text{pos},i}, C_{\text{neg},i})\}_{i=1}^{N} \), where \( I_i \) is an image, \( c_{\text{pos},i} \) is its corresponding positive logical caption, and \( C_{\text{neg},i} = \{ c_{\text{neg},i,1}, c_{\text{neg},i,2}, c_{\text{neg},i,3} \} \) are the three hard negative captions generated by logical perturbations, we construct a set of four description options \( C_{\text{options},i} = \{ c_{\text{pos},i} \} \cup C_{\text{neg},i} \). The model is tasked with identifying the correct caption \( c_{\text{pos},i} \) among these four options for each image \( I_i \).
We compute the cosine similarity between the image feature \( \mathbf{v}_i = E_\text{I}(I_i) \). This leads to a set of logits \( \mathbf{l}_i = \{\cos(\mathbf{v}_i, \mathbf{t}_{\text{option},m})\}_{m=1}^{4} \), representing the similarity between the image and each option.
The multiple-choice loss \( L_{\text{MC}} \) is then computed by applying a cross-entropy loss over the logits, with the index of \( c_{\text{pos},i} \) as the ground-truth target:
\[
L_{\text{MC}} = -\log \frac{\exp(\mathbf{l}_{i, \text{pos}})}{\sum_{m=1}^{4} \exp(\mathbf{l}_{i,m})},
\]
where \( \mathbf{l}_{i, \text{pos}} \) denotes the logit corresponding to the positive caption \( c_{\text{pos},i} \). This encourages the model to assign higher scores to logically correct captions and lower scores to incorrect ones, fostering fine-grained logical discrimination.

\noindent \textbf{Logical Structure-Aware Objective.} 
The aim of this objective is to explicitly guide VLMs to identify specific logical structures within a given text description, further enhancing the model's attention to logical relations rather than merely focusing on surface-level semantic matching.
For each caption \( C_i \) in a batch, we obtain its text encoder feature \( \mathbf{t}_i = E_\text{T}(C_i) \). We then feed \( \mathbf{t}_i \) into a logical classifier \( F_\text{Logic} \), which outputs a vector of predicted scores for each logical category in \( \mathcal{L} \). The classifier assigns a score for each category, indicating the likelihood that caption \( C_i \) belongs to that category. Since a caption \( C_i \) can contain multiple logical categories, its ground-truth label is a multi-hot encoded vector \( y_i \in \{0, 1\}^k \), where \( y_{i,j} = 1 \) if \( C_i \) contains logical category \( L_j \) and \( 0 \) otherwise. We train this logical classifier using binary cross-entropy loss $L_\text{Logic}$. This loss explicitly encourages the model to accurately predict all logical categories present in the text description, thereby enhancing its direct comprehension of logical structures.

\noindent \textbf{Total Optimization Objective.} 
The total optimization objective is a weighted sum of these three components:
\[
L_{\text{Total}} = \alpha L_{\text{CLIP}} + \beta L_{\text{MC}} + \gamma L_\text{Logic},
\]
where $\alpha$, $\beta$, and $\gamma$ are hyperparameters that control the contribution of each loss term. This composite objective ensures that LogicCLIP maintains general vision-language alignment, accurately distinguishes subtle logical differences, and explicitly understands various logical relations through direct classification.

\section{V. Experiments}

\begin{table*}[h]
\centering
\resizebox{\textwidth}{!}{
\begin{tabular}{l|cccccccc|cccc}
\toprule
\multirow{3}{*}{Model}  & \multicolumn{8}{c}{\textbf{LogicBench}} & \multicolumn{4}{c}{\textbf{General Benchmark}} \\ 
  &     \multicolumn{3}{c}{\textit{Image}}  & \multicolumn{3}{c}{\textit{Video}} & \textit{Anomoly} & \textit{Medicine} & \multicolumn{2}{c}{\textit{COCO}} & \multicolumn{2}{c}{\textit{Flickr30K}}  \\ 
  &      \textbf{MCQ} & \textbf{R@1} & \textbf{R@5} & \textbf{MCQ} & \textbf{R@1} & \textbf{R@5} & \textbf{MCQ} & \textbf{MCQ} & \textbf{R@1} & \textbf{R@5} & \textbf{R@1} & \textbf{R@5} \\
\midrule
Human   & 96.32    & -     & -     & 93.97      & -     & -& 94.23     & 87.00     & -      & -      & -& -\\
\midrule
LaCLIP  & 42.28 & 28.80 & 52.47 & 42.55 & 44.40 & 69.73 & 35.14 & 13.50 & 31.67 & 56.12 & 57.58 & 82.14 \\
LaionCLIP & 50.81 & 36.41 & 61.73 & 46.32 & 56.40 & 78.53 & 30.92 & 15.40 & 39.37 & 65.42 & 66.76 & 88.38 \\
DatacompCLIP & 36.60 & 8.94 & 23.34 & 35.47 & 22.53 & 44.80 & 32.93 & 7.40 & 10.98 & 27.02 & 18.10 & 40.36 \\
TripletCLIP-CC3M & 44.37 & 3.50 & 10.39 & 39.81 & 7.33 & 20.67 & 16.62 & 26.20 & 3.59 & 11.29 & 8.72 & 22.08 \\
TripletCLIP-CC12M & 56.32 & 10.18 & 26.15 & 48.49 & 21.33 & 44.93 & 15.66 & 17.50 & 11.25 & 28.20 & 25.36 & 51.76 \\
MetaCLIP & 54.05 & 33.06 & 58.26 & 44.34 & 52.80 & 76.80 & 25.85 & 17.00 & 36.62 & 62.48 & 63.84 & 86.14 \\
ILCLIP  & 54.55 & 9.88 & 25.21 & 30.94 & 14.80 & 33.33 & 24.00 & 17.30 & 12.35 & 30.27 & 20.52 & 43.36 \\ 
ConCLIP & 38.58 & 23.00 & 45.09 & 42.08 & 44.53 & 68.67 & 30.22 & 8.30 & 27.67 & 52.26 & 56.74 & 82.68 \\ 
NegFull & 60.04 & 27.00 & 50.98 & 55.85 & 45.60 & 71.60 & 31.22 & 17.90 & 30.60 & 55.49 & 57.68 & 82.54 \\
NegCLIP    & 55.27 & 39.38 & 67.06 & 53.96 & 59.07 & 82.00 & 32.03    & 21.30  & 41.52 & 68.43 & 67.44 & 89.50  \\
OpenAI CLIP-B   & 38.56 & 27.88 & 50.85 & 35.57 & 49.60 & 72.40 & 28.36 & 16.50 & 30.44 & 55.97 & 58.78 & 83.54  \\
OpenAI CLIP-L   & 36.90 & 34.24 & 58.38 & 35.66 & 53.73 & 76.40 & 24.75 & 12.80 & 36.52 & 61.06 & 65.00 & 87.26   \\
\midrule
LogicCLIP-NegFull (Ours) & 79.34 & 39.85 & 67.64 & 70.85 & 57.60 & 82.93 & \underline{57.83} & \underline{37.60} & 42.77 & 69.18 & 67.20 & 89.44 \\
LogicCLIP-Neg (Ours) & \textbf{85.69} & \underline{42.30} & \underline{69.50} & \textbf{82.55} & \textbf{59.33} & 82.67 & 56.38  & \textbf{46.60}  & \underline{44.38}  & \underline{71.28}  & \underline{69.84}   & \underline{90.58}\\
LogicCLIP-B (Ours) & 81.91 & 39.82 & 67.00 & 77.92  & 57.60 & \textbf{83.33}   & 45.33 & 33.80 & 42.54  & 69.74  & 68.78  & 89.66  \\
LogicCLIP-L (Ours) & \underline{83.93} & \textbf{44.28} & \textbf{71.46} & \underline{79.53}  & \underline{58.53} & \underline{83.07} & \textbf{62.30} & 35.90 & \textbf{45.27}  & \textbf{71.54}  & \textbf{71.98}  & \textbf{91.76}  \\
\bottomrule
\end{tabular}
}
\caption{Performance comparison of different models on LogicBench and general benchmark datasets. \textbf{Bold} and \underline{underline} show the best and second-best performances.}
\label{tab1}
\end{table*}

\subsection{Experimental Settings}

\noindent \textbf{Evaluated Models.}
We conduct a comprehensive evaluation of state-of-the-art VLMs on LogicBench and general benchmarks, including OpenAI's native CLIP (base-32 and large-14 versions) \cite{radford2021learning}, LaCLIP \cite{fan2023improving}, LaionCLIP \cite{schuhmann2022laion}, DatacompCLIP \cite{gadre2023datacomp}, NegCLIP \cite{yuksekgonul2023when}, TripletCLIP \cite{patel2024tripletclip}, MetaCLIP \cite{xu2024demystifying}, ILCLIP \cite{Zheng_2024_CVPR}, ConCLIP \cite{singh2025learning}, and NegFull \cite{alhamoud2025vision}. 
Additionally, we include a set of human evaluation experiments by randomly selecting 10\% of the samples from LogicBench. Independent human evaluators answered each question and provided assessments, serving as a reference for human performance.

\noindent \textbf{Benchmarks \& Metrics.}
In addition to LogicBench, we also evaluate the performance of VLMs on general benchmarks to comprehensively assess model capabilities. Following \cite{zhang2024long,wang2025spatialclip}, we use COCO \cite{lin2014microsoft} and Flickr30k \cite{plummer2015flickr30k} for text-image retrieval. For MCQ tasks, we report accuracy. 
For retrieval tasks, we report Recall@1 and Recall@5.

\noindent \textbf{Training Settings.}
To evaluate the efficacy of our LogicCLIP framework, we fine-tune four representative pretrained VLMs: OpenAI CLIP-B, OpenAI CLIP-L, NegCLIP, and NegFull. We then compare the fine-tuned models with their respective base versions on both LogicBench and general benchmarks. All models are fine-tuned for 16 epochs with a 1,000-step linear warmup. We employ the AdamW optimizer with a weight decay of 0.2. Batch sizes are set to 256 for OpenAI CLIP-B and NegCLIP. Due to memory constraints, OpenAI CLIP-L is fine-tuned with a reduced batch size of 64. Experiments are implemented using PyTorch and conducted on a single NVIDIA A100-80G GPU. The hyperparameters $\alpha$, $\beta$, and $\gamma$ are set to 4, 2, and 1, respectively.

\begin{figure*}[h]
\centering
	\includegraphics[trim=0 5 0 5, clip=true, width=0.90\textwidth]{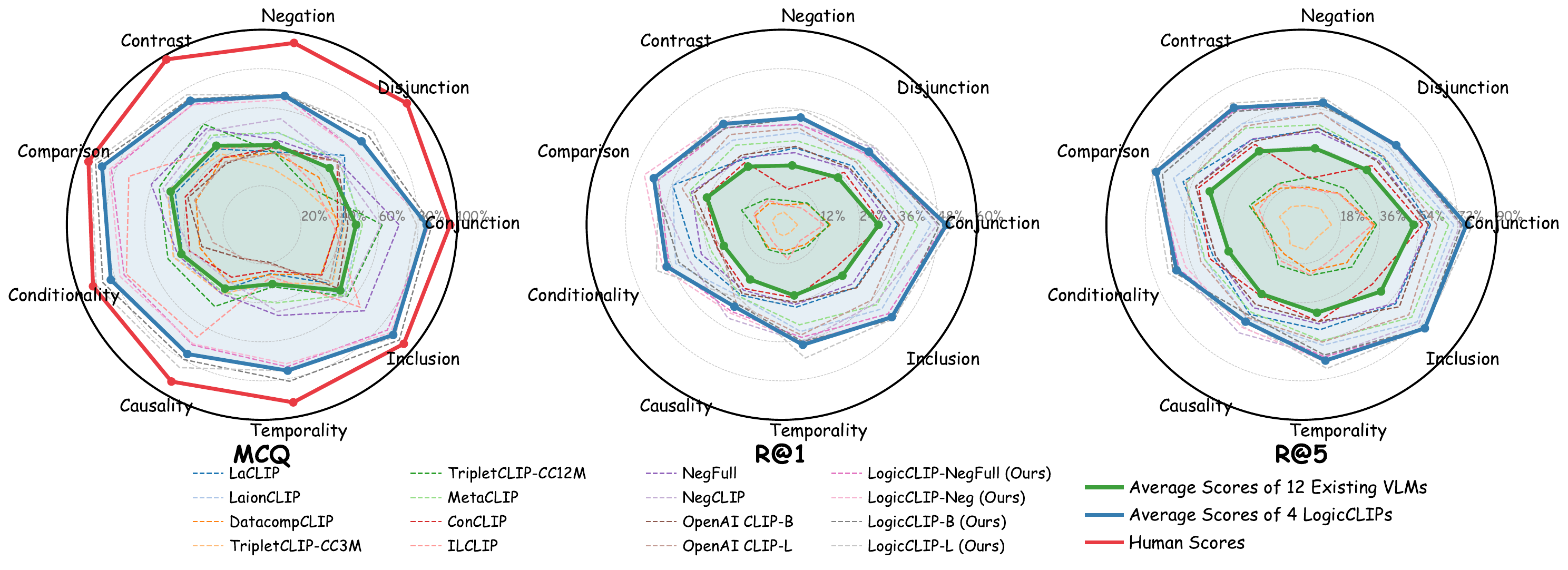}
	\caption{Performance comparison across different logical structures. \textit{See Supp. Mat. C for more detailed results.}}
 \label{fig4}
\end{figure*}

\subsection{Results \& Insights}
\noindent \textbf{1. Logic: A major challenge faced by current VLMs.}
To begin, we evaluate the performance of various VLMs on LogicBench, as shown in Table \ref{tab1}. These results reveal significant shortcomings of current VLMs in logical understanding. 
Even for relatively simple image scenes, models like LaionCLIP and MetaCLIP, which perform well on general benchmarks, achieve only 50.81\% and 54.05\% on the MCQ task, respectively. The strongest baseline, NegFull, scores 60.04\% on MCQ but only 27.00\% on R@1. Furthermore, models show even weaker performance in more complex scenarios such as video, anomaly detection, and medical diagnostics, with MCQ scores dropping significantly. Other competitive pre-trained models, such as LaCLIP (42.28\% on image MCQ), LaionCLIP (50.81\% on image MCQ), and the original OpenAI CLIP-B (38.56\% on image MCQ), achieve more moderate scores. In contrast, human evaluators perform exceptionally well across all four LogicBench scenarios, with scores surpassing 90\% in image, video, and anomaly tasks. 
These results collectively underscore that the current VLMs are inadequate for handling logical relations.

\noindent \textbf{2. LogicCLIP achieves remarkable improvements on logical tasks.} 
In contrast, our LogicCLIP consistently demonstrates superior performance across all LogicBench scenarios, significantly outperforming its original versions. LogicCLIP-B and LogicCLIP-L show a significant boost in performance on LogicBench, particularly in MCQ tasks and R@1 across all application scenarios. For example, LogicCLIP-B achieves an MCQ score of 81.91\%, far surpassing the OpenAI CLIP-B score of 38.56\%. Similarly, LogicCLIP-L elevates Image MCQ from 36.90\% to 83.93\%, and LogicCLIP-Neg achieves the highest Image MCQ score at 85.69\% compared to NegCLIP's 55.27\%. These substantial improvements extend across other logical tasks within LogicBench: LogicCLIP-B improves Video MCQ from 35.57\% to 77.92\%, Anomaly MCQ from 28.36\% to 45.33\%, and Medicine MCQ from 16.50\% to 33.80\%. These results demonstrate that the fine-tuning process, which incorporates logic-aware training, has significantly enhanced the VLMs' ability to reason logically.

\noindent \textbf{3. Logic-aware training also improves performance on general benchmarks.} 
On general benchmarks, our model not only maintains the performance of its base model but often achieves significant improvements. For instance, LogicCLIP-B improves the R@1 accuracy on COCO from 30.44\% (OpenAI CLIP-B) to 42.54\%, and on Flickr30K, the R@1 accuracy increases from 58.78\% to 68.78\%. These results indicate that LogicCLIP enhances the model's ability to capture deeper logical relationships, prompting it to build richer, more nuanced image and text representations. As a result, the overall quality of image-text alignment is improved. This enhanced understanding likely contributes to better generalization on general retrieval tasks.
Our findings suggest that LogicCLIP has the potential to guide the development of more robust and powerful vision-language foundation models.

\noindent \textbf{4. Broad generalization across unseen logical domains.} 
It is important to emphasize the remarkable generalization capabilities of LogicCLIP. While our training data is derived solely from daily images, the logic-aware abilities acquired by LogicCLIP effectively transfer to unseen, specialized domains within LogicBench. For instance, LogicCLIP-B boosts Video MCQ from 35.57\% to an impressive 77.92\%. LogicCLIP-L boosts Anomaly MCQ from 24.75\% to 62.30\%, and LogicCLIP-Neg improves Medicine MCQ from 21.30\% to 46.60\%. This demonstrates that LogicCLIP learns a fundamental understanding of logical structures that is transferable across diverse visual contexts.

\noindent \textbf{5. Varying difficulty across logical categories.}
Beyond overall performance, analyzing results across specific logical structures offers deeper insights into model capabilities and limitations. 
Figure \ref{fig4} details the MCQ, R@1, and R@5 scores for 9 distinct logical categories. 
Our analysis reveals a clear spectrum of difficulty for current baseline VLMs in discerning different logical relationships.
\textit{\textbf{(1)} Temporality, Causality, Conditionality, and Negation prove to be the most challenging}, with average baseline MCQ scores of 31.18\%, 35.15\%, 41.23\%, and 41.47\%, respectively. For example, OpenAI CLIP-B/L consistently scores below 20\% MCQ on Temporality and Causality, highlighting its weakness in handling nuanced temporal and causal relationships.
\textit{\textbf{(2)} Disjunction, Contrast, and Comparison present moderate challenges}, with average MCQ scores of 46.28\%, 46.97\%, and 47.71\%. Baselines score in the 40-60\% range, suggesting partial understanding but substantial room for improvement.
\textit{\textbf{(3)} Conjunction and Inclusion are comparatively less challenging for some strong baselines}, with average MCQ scores of 49.26\% and 51.18\%, respectively. LogicCLIP-NegFull achieves 70.30\% on Conjunction and 68.60\% on Inclusion, though standard CLIP variants show only moderate performance in these areas.
Our LogicCLIP consistently demonstrates superior performance across all nine logical categories, dramatically overcoming the limitations of baseline models. In challenging categories like Temporality, Causality, and Conditionality, LogicCLIP's MCQ scores frequently leap from baseline ~20-40\% to over 80-90\%. It also excels in other logical structures, including Comparison, Conjunction, and Inclusion, regularly achieving over 90\% MCQ accuracy. This comprehensive improvement across the spectrum of logical complexities proves LogicCLIP effectively instills robust, granular logical understanding in VLMs.
\textit{More ablations, qualitative analyses, and visualizations are in Supp. Mat. C.}



\section{VI. Conclusion}
In this paper, we tackle the critical challenge of logical understanding in VLMs, a crucial yet underexplored capability for their reliable deployment. 
We introduce LogicBench, a comprehensive benchmark designed to assess the logical reasoning capabilities of VLMs. Our evaluation reveals significant logical blindspots in current VLMs, particularly in tasks involving Temporality, Causality, Conditionality, and Negation.
We further analyze that these blindspots stem from a lack of explicit consideration of logic during both data collection and optimization objectives.
To address these limitations, we propose LogicCLIP, a novel training framework that enhances VLMs' logical sensitivity by integrating a large-scale hard negative sample generation pipeline and a logic-aware contrastive learning strategy, both focused on improving logical understanding.
Extensive experiments demonstrate that LogicCLIP not only outperforms state-of-the-art models on LogicBench but also retains competitive performance on general vision-language benchmarks. Additionally, LogicCLIP generalizes effectively across diverse domains such as video, anomaly detection, and medical diagnostics. Our work highlights the critical role of explicit logical training in VLMs and sets the stage for building more robust and reliable models for real-world applications.

\bibliography{aaai2026}

\section{A. More Details about LogicBench}
\label{SecA}


\subsection{A.1 Dataset Statistics}
\label{SedA1}
LogicBench evaluates the logical understanding of Vision-Language Models (VLMs) across four distinct real-world scenarios: natural scene images, videos, anomaly detection, and medical diagnostics. These scenarios represent diverse contexts in which strong logical comprehension is essential for effective VLM performance. The statistics for each scenario are as follows:

\begin{itemize}
    \item \textbf{LogicBench-Image:}
This scenario includes 8,788 multiple-choice questions (MCQ) and 8,788 retrieval tasks, sourced from the MSCOCO \cite{lin2014microsoft} and CC12M \cite{changpinyo2021conceptual} validation sets. These datasets provide highly diverse images captured from everyday environments, testing the model’s ability to reason over natural scene images. Each MCQ has four options by default.
    \item \textbf{LogicBench-Video:}
This scenario includes 1,060 MCQs and 1,060 retrieval tasks, sourced from the MSRVTT dataset \cite{xu2016msr}. It features popular video clips from a commercial video search engine, testing the model's capability to understand logical relationships in dynamic video content. Each MCQ has four options by default.
    \item \textbf{LogicBench-Anomaly:}
Featuring 1,992 MCQs, this scenario is based on the DADA-2000 dataset \cite{fang2019dada}. It involves identifying traffic anomalies and accidents, requiring the model to use logical reasoning to detect unusual or hazardous events. Each MCQ has four options by default.
    \item \textbf{LogicBench-Medicine:}
This scenario includes 1,000 MCQs, derived from the Open-i dataset \cite{demner2015preparing}. It contains radiology reports alongside medical images, where the model is required to interpret complex logical relationships, such as negation and causality, in medical findings. To more closely mimic real-world diagnostic processes, we've raised the difficulty in medical scenarios by setting each MCQ to five options by default.
\end{itemize}

\subsection{A.2 More Details on Data Annotation}
\label{SedA2}
\noindent \textbf{Logical Definitions.}
In LogicBench, we systematically define 9 fundamental logical relationships, including conjunction, disjunction, negation, contrast, comparison, condition, causality, temporality, and inclusion. For each logical category, we provide detailed definitions, typical keywords, and additional example expressions, as shown in Figure \ref{fig1}. These definitions enable precise identification of logical relations within vision-language pairs.

\begin{figure*}[h]
\centering
\begin{tcolorbox}[colframe=blue!60!black, colback=blue!10!white, coltitle=white, title=Logical Categories in LogicBench, width=\textwidth, sharp corners=south]
\begin{itemize}
    \item \textbf{Conjunction:} Refers to the combination or co-occurrence of two or more entities.
    \begin{itemize}
        \item \textbf{Typical Keywords:} ``and'', ``both'', ``as well as'', ``not only...but also'', ``together with''
        \item \textbf{Example Expression:} \textit{“Both the cat and the dog are in the house.”}
    \end{itemize}

    \item \textbf{Disjunction:} Refers to a choice relationship, selecting one option from multiple choices.
    \begin{itemize}
        \item \textbf{Typical Keywords:} ``or'', ``either...or'', ``neither...nor'', ``otherwise''
        \item \textbf{Example Expression:} \textit{“You can have either tea or coffee.”}
    \end{itemize}

    \item \textbf{Negation:} Refers to the denial of a certain situation or entity.
    \begin{itemize}
        \item \textbf{Typical Keywords:} ``not'', ``no'', ``never'', ``without'', ``neither...nor''
        \item \textbf{Example Expression:} \textit{“There are no apples in the basket.”}
    \end{itemize}

    \item \textbf{Contrast:} Refers to the opposition or contrast between two entities or states.
    \begin{itemize}
        \item \textbf{Typical Keywords:} ``but'', ``although'', ``however'', ``whereas'', ``while'', ``in contrast''
        \item \textbf{Example Expression:} \textit{“Although it rained, we still went for a walk.”}
    \end{itemize}

    \item \textbf{Comparison:} Refers to the comparison of similarities or differences between entities.
    \begin{itemize}
        \item \textbf{Typical Keywords:} ``more than'', ``less than'', ``as...as'', ``not as...as'', ``most'', ``least'', ``better than''
        \item \textbf{Example Expression:} \textit{“The mountain is higher than the hill.”}
    \end{itemize}

    \item \textbf{Condition:} Refers to the condition under which an event occurs.
    \begin{itemize}
        \item \textbf{Typical Keywords:} ``if'', ``unless'', ``provided that'', ``only if'', ``in case''
        \item \textbf{Example Expression:} \textit{“If it rains, we will stay indoors.”}
    \end{itemize}

    \item \textbf{Causality:} Refers to the cause-effect relationship between events.
    \begin{itemize}
        \item \textbf{Typical Keywords:} ``because'', ``so'', ``due to'', ``as a result'', ``since''
        \item \textbf{Example Expression:} \textit{“The ground is wet because it rained.”}
    \end{itemize}

    \item \textbf{Temporality:} Refers to the chronological order of events.
    \begin{itemize}
        \item \textbf{Typical Keywords:} ``after'', ``before'', ``when'', ``while'', ``then'', ``as soon as''
        \item \textbf{Example Expression:} \textit{“We went for a walk after lunch.”}
    \end{itemize}

    \item \textbf{Inclusion:} Refers to whether something includes or excludes other parts.
    \begin{itemize}
        \item \textbf{Typical Keywords:} ``including'', ``except for'', ``such as'', ``without''
        \item \textbf{Example Expression:} \textit{“The package includes a book and a pen.”}
    \end{itemize}
\end{itemize}
\end{tcolorbox}
\caption{Nine logical categories in LogicBench and their definitions, typical keywords, and example expressions.} 
\label{fig1}
\end{figure*}

\noindent \textbf{Positive Sample Selection.}
We leverage existing human-annotated vision-language datasets and filter captions containing logical structures to create positive samples. Using spaCy and regular expression scripts for syntactic analysis, we first independently scan sentences and match predefined rules to identify common logical structures. If a structure is detected, it is added to a candidate set. SpaCy then performs syntactic parsing to further examine the relationships and enhance the classification of logical categories based on dependency relations. This method allows us to efficiently process large datasets, identifying samples with embedded logical relations. Note that the selected samples are naturally annotated, ensuring authenticity and practical relevance, without artificially introducing abrupt logical structures.

\noindent \textbf{Negative Sample Generation.}
To generate diverse and challenging negative samples, we utilize multiple widely validated large language models (LLMs), including Qwen 2.5-max, DeepSeek-V3, Gemini-2.5-pro, GPT-4.1, and LLaMA 3.3 70B. This multi-LLM approach mitigates biases that could arise from relying on a single model. For LogicBench-Image, LogicBench-Video, LogicBench-Anomaly, and LogicBench-Medicine, we use specific instruction templates (shown in Figures \ref{fig2}-\ref{fig5}) to guide the generation of negative samples. 
For images, videos, and anomaly scenarios, we provide an original, correct caption containing a specific logical structure. The task of LLMs is to create three logically incorrect captions by perturbing the given logical structure. These negative samples must be fluent, plausible, and structurally similar to the original, while introducing logical errors to test the model's ability to identify subtle logical inconsistencies.
For medical scenarios, we provide an authentic diagnostic report, and LLMs generate both correct and logically perturbed options for pathology reports. Perturbations include strategies such as negation flip, conjunction trap, disjunction confusion, and causal misalignment. The goal is to assess the model's ability to discern subtle logical differences within medical contexts.
\textit{After generating both positive and negative samples, human experts review and select the accurate and reliable ones to ensure logical consistency and high quality.}

The dataset is then finalized to form LogicBench, a large-scale, high-quality benchmark that not only meets the scale requirements but also incorporates sufficient logical complexity, naturalness, and diversity. This comprehensive dataset effectively diagnoses current VLMs' logical capabilities, providing a valuable resource for further research and model improvements.

\begin{figure*}[ht!]
\begin{tcolorbox}[colback=gray!5!white, colframe=gray!95!black, title=Instruction Templates for the LogicBench-Image]
You are given a caption that describes an image and contains a \{logic\_type\} logical structure.  
Your task is to generate THREE hard negative captions that:  

- Are fluent and grammatically correct.  

- Appear plausible and similar in structure.  

- Contain incorrect or conflicting \{logic\_type\} logic.  

Caption: ``{caption}"  

Image ID: {image\_id}  

Please output only the THREE hard negative captions as a numbered list, one per line:  

1. $<$hard negative caption$>$

2. $<$hard negative caption$>$

3. $<$hard negative caption$>$

\textit{Placeholder Explanation:}  
\{logic\_type\} refers to one of the following logical structures: conjunction, disjunction, negation, contrast, comparison, condition, causality, temporality, and inclusion.  
\end{tcolorbox}
\caption{Instruction Templates for LogicBench-Image.}
\label{fig2}
\end{figure*}

\begin{figure*}[ht!]
\begin{tcolorbox}[colback=gray!5!white, colframe=gray!95!black, title=Instruction Templates for the LogicBench-Video]
You are given a caption that describes a video and contains a \{logic\_type\} logical structure.  
Your task is to generate THREE hard negative captions that:  

- Are fluent and grammatically correct.  

- Appear plausible and similar in structure.  

- Contain incorrect or conflicting \{logic\_type\} logic.  

Caption: ``{caption}"  

Video ID: {video\_id}  

Please output only the THREE hard negative captions as a numbered list, one per line:  

1. $<$hard negative caption$>$

2. $<$hard negative caption$>$

3. $<$hard negative caption$>$

\textit{Placeholder Explanation:}  
\{logic\_type\} refers to one of the following logical structures: conjunction, disjunction, negation, contrast, comparison, condition, causality, temporality, and inclusion.  
\end{tcolorbox}
\caption{Instruction Templates for LogicBench-Video.}
\label{fig3}
\end{figure*}

\begin{figure*}[ht!]
\begin{tcolorbox}[colback=gray!5!white, colframe=gray!95!black, title=Instruction Templates for the LogicBench-Anomaly]
You are given a caption that describes a abnormal video and contains a \{logic\_type\} logical structure.  
Your task is to generate THREE hard negative captions that:  

- Are fluent and grammatically correct.  

- Appear plausible and similar in structure.  

- Contain incorrect or conflicting \{logic\_type\} logic.  

Caption: ``{caption}"  

Video ID: {video\_id}  

Please output only the THREE hard negative captions as a numbered list, one per line:  

1. $<$hard negative caption$>$

2. $<$hard negative caption$>$

3. $<$hard negative caption$>$

\textit{Placeholder Explanation:}  
\{logic\_type\} refers to one of the following logical structures: conjunction, disjunction, negation, contrast, comparison, condition, causality, temporality, and inclusion.  
\end{tcolorbox}
\caption{Instruction Templates for LogicBench-Anomaly.}
\label{fig4}
\end{figure*}

\begin{figure*}[ht!]
\begin{tcolorbox}[colback=gray!5!white, colframe=gray!95!black, title=Instruction Templates for the LogicBench-Medicine]
You are given a caption that describes a pathology report as follows: \{pathology report\}

Please follow the instructions below:

1. Extract positive findings (statements without negation) and negative findings (statements with negation, e.g., ``no pleural effusion").

2. Construct ONE correct option using this rule-based format:  
   ``Because $<$positive finding$>$ and $<$negative finding$>$, the impression is $<$IMPRESSION$>$."

3. Generate FOUR hard negative options using logical perturbations:

   - Negation Flip: Reverse negation (e.g., ``no effusion" → ``effusion present").
   
   - Conjunction Trap: Combine correct and incorrect statements with ``and".
   
   - Disjunction Confusion: Use ``or" with one correct and one incorrect clause.
   
   - Causal Misalignment: Disrupt the cause-effect chain using ``because", ``so", or ``therefore".

4. Return exactly five options, labeled 1 to 5. Only the first option should be correct.  
Please use the following exact format:  

1. $<$Correct caption$>$  

2. $<$Hard negative caption$>$  

3. $<$Hard negative caption$>$  

4. $<$Hard negative caption$>$  

5. $<$Hard negative caption$>$  

\end{tcolorbox}
\caption{Instruction Templates for LogicBench-Medicine.}
\label{fig5}
\end{figure*}

\subsection{A.3 Visualizing the LogicBench Evaluation Tasks}
\label{SedA3}
In this section, we provide visualizations of example cases from LogicBench across the image, video, anomaly, and medical scenarios for reference. As the vision-language pairs for the retrieval task are identical to those used in the MCQ task, we focus our visualizations on the MCQ task, presented in Figures \ref{fig6}, \ref{fig7}, \ref{fig8}, and \ref{fig9}.

\section{B. More Details of LogicCLIP}
\subsection{B.1 Training Data}
Our logic-aware training data builds upon the widely used MSCOCO dataset \cite{lin2014microsoft}, which provides human-annotated captions for images from everyday scenes. We augment this dataset to train LogicCLIP and enhance its ability to understand and process complex logical structures. The training data consists of 475,624 logical caption samples, designed to capture various logical categories and relationships present in everyday visual and textual descriptions.
The data is carefully curated with 25\% positive captions, which contain explicitly defined logical structures, and 75\% negative captions, generated through logical perturbations. This allows the model to effectively learn to identify not only correct logical structures but also subtle distinctions between logically correct and incorrect captions. 

\subsection{B.2 Model Variants}
To evaluate the performance of LogicCLIP, we fine-tune it on four representative state-of-the-art VLMs that serve as the baseline for comparison:
\begin{itemize}
\item \textbf{OpenAI CLIP-B} \cite{radford2021learning}: The base model of CLIP, available at \url{https://huggingface.co/openai/clip-vit-base-patch32}, offers a standard architecture for fine-tuning across various downstream tasks. It is commonly used as a baseline in vision-language model evaluation.
\item \textbf{OpenAI CLIP-L} \cite{radford2021learning}: The large variant of CLIP, accessible at \url{https://huggingface.co/openai/clip-vit-large-patch14}, has a larger capacity compared to the base model and is typically expected to perform better in tasks requiring more expressive power.
\item \textbf{NegFull} \cite{alhamoud2025vision}: This model, which specifically addresses logic in vision-language tasks, is available along with its code at \url{https://github.com/m1k2zoo/negbench}. 
\item \textbf{NegCLIP} \cite{yuksekgonul2023when}: NegCLIP, available at \url{https://github.com/mertyg/vision-language-models-are-bows}, focuses on incorporating composition-aware learning into CLIP. 
\end{itemize}

\section{C. More Details of Experiments}
\label{SecC}
\subsection{C.1 Baselines}
\label{SecC1}
In the main paper, we conduct a comprehensive evaluation of current state-of-the-art VLMs on LogicBench and general benchmarks. These include OpenAI's native CLIP (base-32 and large-14 versions), along with several other prominent models: LaCLIP, LaionCLIP, DatacompCLIP, TripletCLIP, MetaCLIP, ConCLIP, ILCLIP, NegFull, and NegCLIP. For reproducibility, the details on how to acquire or access these baseline models are provided below:
\begin{itemize}
    \item \textbf{OpenAI CLIP-B} \cite{radford2021learning}: The model is available at \url{https://huggingface.co/openai/clip-vit-base-patch32}.
    \item \textbf{OpenAI CLIP-L} \cite{radford2021learning}: The model is available at \url{https://huggingface.co/openai/clip-vit-large-patch14}.
    \item \textbf{LaCLIP} \cite{fan2023improving}:  The model and code are available at \url{https://github.com/LijieFan/LaCLIP}.
    \item \textbf{LaionCLIP} \cite{schuhmann2022laion}:  The model is available at \url{https://huggingface.co/laion/CLIP-ViT-B-32-laion2B-s34B-b79K}.
    \item \textbf{DatacompCLIP} \cite{gadre2023datacomp}: The model is available at \url{https://huggingface.co/laion/CLIP-ViT-B-32-DataComp.M-s128M-b4K}.
    \item \textbf{TripletCLIP-CC3M} \cite{patel2024tripletclip}: This variant is accessible via \url{https://huggingface.co/TripletCLIP/CC3M_TripletCLIP_ViTB12}.
    \item \textbf{TripletCLIP-CC12M} \cite{patel2024tripletclip}: This variant is accessible via \url{https://huggingface.co/TripletCLIP/CC12M_TripletCLIP_ViTB12}.
    \item \textbf{MetaCLIP} \cite{xu2024demystifying}: The model and code are available at \url{https://github.com/facebookresearch/MetaCLIP}.
    \item \textbf{ConCLIP} \cite{singh2025learning}: The model and code are available at \url{https://github.com/jaisidhsingh/CoN-CLIP}.
    \item \textbf{ILCLIP} \cite{Zheng_2024_CVPR}: The model and code are available at \url{https://github.com/hellomuffin/iterated-learning-for-vlm}.
    \item \textbf{NegFull} \cite{alhamoud2025vision}: The model and code are available at \url{https://github.com/m1k2zoo/negbench}. 
    \item \textbf{NegCLIP} \cite{yuksekgonul2023when}: The model and code are available at \url{https://github.com/mertyg/vision-language-models-are-bows}. 
\end{itemize}


\begin{table*}[t]
\centering
\resizebox{\textwidth}{!}{
\begin{tabular}{c|ccc|ccc|c|c|cc|cc}
\toprule
\multirow{3}{*}{Model}  & \multicolumn{8}{c|}{\textbf{LogicBench}} & \multicolumn{4}{c}{\textbf{General Benchmark}} \\ 
  & \multicolumn{3}{c}{\textit{Image}}  & \multicolumn{3}{c}{\textit{Video}} & \textit{Anomaly} & \textit{Medicine} & \multicolumn{2}{c}{\textit{COCO}} & \multicolumn{2}{c}{\textit{Flickr30K}}  \\ 
  & \textbf{MCQ} & \textbf{R@1} & \textbf{R@5} & \textbf{MCQ} & \textbf{R@1} & \textbf{R@5} & \textbf{MCQ} & \textbf{MCQ} & \textbf{R@1} & \textbf{R@5} & \textbf{R@1} & \textbf{R@5} \\
\midrule
OpenAI CLIP-B & 38.56 & 27.88 & 50.85 & 35.57 & 49.60 & 72.40 & 28.36 & 16.50 & 30.44 & 55.97 & 58.78 & 83.54 \\ \hline
Variant 1 & 91.29 & 0.23 & 1.04 & 94.43 & 0.40 & 2.13 & 83.79 & 20.80 & 0.16 & 0.90 & 0.10 & 2.22 \\
Variant 2 & 71.03 & 0.17 & 0.56 & 70.38 & 0.27 & 1.07 & 74.10 & 0.50 & 0.02 & 0.11 & 0.06 & 0.42 \\
Variant 3 & 60.18 & 33.51 & 59.66 & 52.17 & 47.73 & 75.87 & 26.15 & 16.30 & 35.75 & 62.61 & 57.76 & 83.28 \\
Variant 4 & 77.67 & 39.74 & 66.34 & 74.15 & 56.67 & 79.73 & 41.82 & 24.00 & 41.47 & 68.35 & 66.68 & 88.88 \\ \hline
LogicCLIP-B & 81.91 & 39.82 & 67.00 & 77.92  & 57.60 & 83.33 & 45.33 & 33.80 & 42.54  & 69.74  & 68.78  & 89.66 \\
\bottomrule
\end{tabular}}
\caption{Performance comparison of different variants on LogicBench and General Benchmark tasks.}
\label{tab:ablation_loss}
\end{table*}

\subsection{C.2 Ablation Studies on Optimization Objective}
To thoroughly understand the contribution of each component within our Logic-Aware Contrastive Learning framework, we conduct an ablation study by training LogicCLIP with various subsets of its proposed loss functions, building upon OpenAI CLIP-B. We evaluate four variations:
\textbf{(1)} Fine-Grained Multiple-Choice Objective ($L_{\text{MC}}$) alone; 
\textbf{(2)} Logical Structure-Aware Objective ($L_{\text{Logic}}$) alone; 
\textbf{(3)} Standard CLIP Contrastive Objective ($L_{\text{CLIP}}$) alone; 
\textbf{(4)} A combination of $L_{\text{CLIP}}$ and $L_{\text{MC}}$, excluding $L_{\text{Logic}}$.
Table \ref{tab:ablation_loss} presents these results, showcasing performance on both LogicBench and general benchmarks.

Our analysis reveals that while individual logical loss components ($L_{\text{MC}}$ and $L_{\text{Logic}}$) are highly effective for logical understanding, they are insufficient for maintaining general vision-language alignment. When trained solely with $L_{\text{MC}}$, the model achieves an impressive Image MCQ score of 0.9129, demonstrating its critical role in fine-grained logical distinction. Similarly, $L_{\text{Logic}}$ alone yields a respectable 0.7103 Image MCQ. However, models trained with only $L_{\text{MC}}$ or $L_{\text{Logic}}$ exhibit extremely poor performance on general benchmarks (e.g., COCO R@1 for $L_{\text{MC}}$ is 0.0016, and for $L_{\text{Logic}}$ is 0.0002), indicating a severe loss of general VLM capabilities. Conversely, training only with $L_{\text{CLIP}}$ (0.6018 Image MCQ, 0.3575 COCO R@1) shows moderate logical understanding but maintains general alignment, similar to the OpenAI CLIP-B baseline (0.3856 Image MCQ, 0.3044 COCO R@1).

In contrast, the combined objectives demonstrate powerful synergistic effects:
\textbf{(1)} Adding $L_{\text{MC}}$ to $L_{\text{CLIP}}$ (Variant 4) significantly boosts logical understanding (Image MCQ jumps to 0.8191) while simultaneously improving general benchmark performance (COCO R@1 increases to 0.4147 from 0.3575 for $L_{\text{CLIP}}$ alone), proving $L_{\text{MC}}$ effectively enhances logical reasoning without compromising general alignment when combined with $L_{\text{CLIP}}$.
\textbf{(2)} The full model, incorporating all three objectives (Full LogicCLIP-B), achieves the best overall performance. It further elevates logical comprehension (Image MCQ reaching 0.8191), providing an additional boost over the $L_{\text{CLIP}}+L_{\text{MC}}$ combination. Critically, this full integration maintains the strong general VLM capabilities (COCO R@1 is 0.4254), demonstrating that $L_{\text{Logic}}$ provides a distinct, valuable contribution to explicit logical categorization without degrading broader VLM performance.

\begin{table}[h]
\centering
\resizebox{0.48\textwidth}{!}{
\begin{tabular}{l|cccc}
\toprule
\textbf{Model} & \textbf{Image} & \textbf{Video} & \textbf{Anomaly} & \textbf{Medicine} \\
\midrule
CLIP-L & 36.90 & 35.66 & 24.75 & 12.80 \\
LLaVA1.5-7b w/ CLIP-L & 63.94 & 49.72 & 48.34 & 26.70 \\
LogicCLIP-L (Ours) & 83.93 & 79.53 & 62.30 & 35.90 \\
\bottomrule
\end{tabular}
}
\caption{LLaVA's performance on LogicBench MCQ tasks compared to CLIP-L and LogicCLIP-L.}
\label{tablella} 
\end{table}

\subsection{C.3 Evaluating LLaVA on LogicBench}
\label{SecC2}
In the main paper, we propose a novel benchmark for evaluating logical understanding and focus primarily on joint-embedding VLMs, especially CLIP-based architectures, which are dominant in multimodal retrieval tasks. We also acknowledge that Multi-Modal Large Language Models (MLLMs) can be relevant. Specifically, MLLMs like LLaVA have paved the way for conversational VLM chatbots. Here, we evaluate LLaVA's performance on the MCQ tasks across all scenarios of LogicBench to understand its logical reasoning capabilities. The results are presented in Table \ref{tablella}, which provides a comparison with our proposed LogicCLIP-L model and the base CLIP-L model.

The instruction-tuned LLaVA 1.5-7B model demonstrates a significant improvement in logical understanding compared to its base CLIP-L model. For instance, LLaVA achieves a 63.94\% MCQ score on the image scenario, a substantial gain over CLIP-L's 36.90\%. It also shows notable improvements in video (49.72\% vs. 35.66\%) and anomaly detection (48.34\% vs. 24.75\%).
A key strength of LLaVA likely stems from its use of a powerful LLM like Vicuna for text encoding. Unlike CLIP, which is primarily pretrained on simple image descriptions, LLMs like Vicuna are trained on diverse text corpora containing various logical structures, enabling LLaVA to better interpret logical statements. However, this approach does not scale well to a large number of candidate choices and is impractical for text-to-image retrieval, where placing all dataset images into the context window is unfeasible. Therefore, advancing models like CLIP remains crucial for real-world multimodal retrieval tasks with logical relations.

\begin{figure*}[t]
\centering
	\includegraphics[trim=0 0 0 0, clip=true, width=0.8\textwidth]{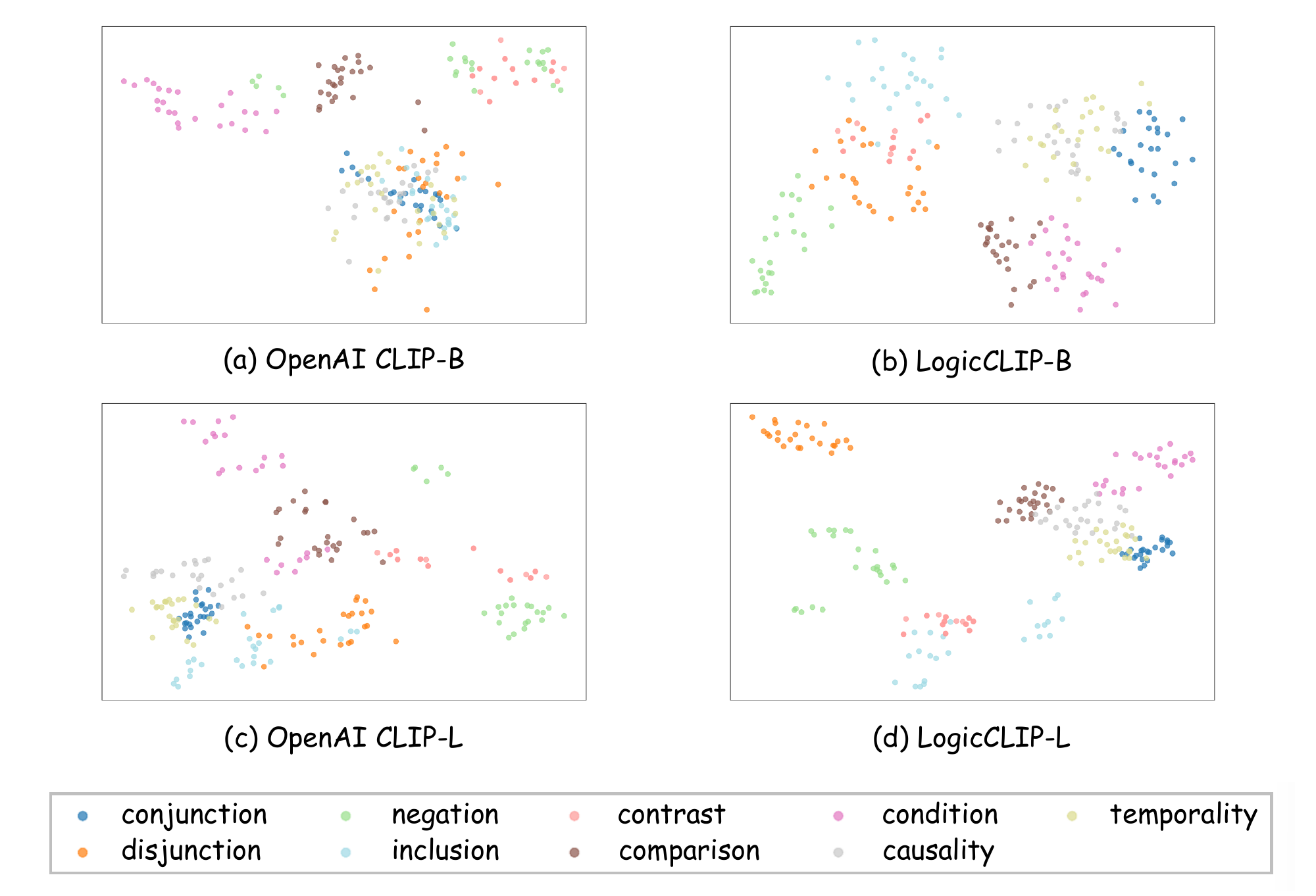}
	\caption{t-SNE visualization of caption embeddings.} 
 \label{fig_emdvis}
\end{figure*}

\subsection{C.4 Feature Distribution Analysis}
We further investigate the model's embedding space by applying different logic-specific templates to a single caption within a particular scenario. Specifically, we begin with the caption:
\textit{``The family enjoyed the colorful frisbee, tasty sandwiches, a comfortable blanket, and playing with their lively dog.''} We then generated 24 variations of this sentence by applying logic-specific templates for each of the 9 logical categories. The resulting sentences had similar constituent components (e.g., all included ``family," ``frisbee," ``blanket," ``sandwiches," and ``dog") but had different meanings due to subtle logical shifts caused by the variation in logical words. Figure \ref{fig_emdvis} shows the t-SNE plots for OpenAI CLIP-B, LogicCLIP-B, OpenAI CLIP-L, and LogicCLIP-L.

As shown in Figure \ref{fig_emdvis}(a)(c), the embeddings from both OpenAI CLIP-B and CLIP-L models exhibit significant overlap between different logical categories. This is likely due to the ``bag-of-words" effect, where the model primarily focuses on individual words and phrases rather than their logical relationships. For example, CLIP-B shows significant overlap in logical categories such as causal, conjunction, disjunction, temporal, and inclusion, indicating its inability to distinguish the semantic changes brought about by different logical words. In contrast, Figure \ref{fig_emdvis}(b)(d) reveals that the LogicCLIP-B and LogicCLIP-L embeddings form well-separated and tightly-knit clusters for each logical template. For instance, conjunction captions form a distinct cluster that is clearly separated from captions expressing disjunction and negation. In LogicCLIP-L, logical categories such as comparison, conditionality, and causality are also embedded into relatively unique regions of the t-SNE space. This clear clustering demonstrates that LogicCLIP effectively enhances the model's understanding of logical reasoning, a capability that is largely absent in the original CLIP models. Notably, this improvement in logical awareness does not come at the expense of general alignment and may even facilitate it (as shown by the General Benchmark evaluation in Table 1 of the main paper).


\subsection{C.5 Detailed Performance Across Logical Categories}
\label{SecC3}


As a supplement to Section 5 in the main paper and to Figure 4, we provide a detailed breakdown of model performance across the nine logical categories introduced in LogicBench. The full results are presented in Tables \ref{tab:logical_comparison3}, \ref{tab:logical_comparison4}, and \ref{tab:logical_comparison5}.

As discussed in the main paper, existing VLMs exhibit significant limitations, with \textbf{Temporality}, \textbf{Causality}, \textbf{Conditionality}, and \textbf{Negation} proving to be the most challenging categories. For instance, baseline models often score below 45\% MCQ, and in the case of OpenAI CLIP-B/L, scores drop to around 20\% on Temporality and Causality. In stark contrast, LogicCLIP variants consistently demonstrate superior performance, effectively overcoming these challenges. Our models achieve dramatic gains in these difficult categories, with MCQ scores frequently leaping from the baseline range of 20-40\% to over 80-90\%. For example, LogicCLIP-B boosts Temporality MCQ to 81.46\% (from 19.46\% for CLIP-B) and LogicCLIP-L achieves 90.42\% MCQ on Conditionality (from 26.61\%). Furthermore, LogicCLIP also excels in the ``easier" categories, regularly achieving near-perfect scores on \textbf{Comparison} (e.g., LogicCLIP-L at 92.93\% MCQ) and \textbf{Inclusion} (e.g., LogicCLIP-B at 91.87\% MCQ). The consistent improvement across all logical complexities, as also reflected in R@1 and R@5 scores, unequivocally proves that LogicCLIP's multi-faceted approach effectively instills a robust, granular logical understanding across the full spectrum of relationships.

\begin{table*}[t]
\centering
\resizebox{0.9\textwidth}{!}{
\begin{tabular}{l|ccc|ccc|ccc}
\toprule
\textbf{Model} & \multicolumn{3}{c|}{\textbf{Conjunction}} & \multicolumn{3}{c|}{\textbf{Disjunction}} & \multicolumn{3}{c}{\textbf{Negation}} \\
\cmidrule(r){2-4} \cmidrule(r){5-7} \cmidrule(r){8-10}
& \textbf{MCQ} & \textbf{R@1} & \textbf{R@5} & \textbf{MCQ} & \textbf{R@1} & \textbf{R@5} & \textbf{MCQ} & \textbf{R@1} & \textbf{R@5} \\
\midrule
LaCLIP & 41.71 & 35.71 & 59.50 & 55.40 & 28.27 & 48.95 & 37.91 & 23.98 & 45.09 \\
LaionCLIP & 50.15 & 45.91 & 70.28 & 53.36 & 34.18 & 52.74 & 47.76 & 28.84 & 52.23 \\
DatacompCLIP & 39.25 & 14.98 & 33.70 & 37.98 & 10.13 & 22.57 & 37.45 & 6.24 & 17.54 \\
TripletCLIP-CC3M & 42.41 & 4.90 & 13.88 & 28.51 & 3.38 & 11.60 & 29.49 & 3.96 & 9.02 \\
TripletCLIP-CC12M & 61.66 & 14.57 & 34.62 & 30.55 & 10.34 & 26.16 & 37.53 & 7.63 & 22.40 \\
MetaCLIP & 51.51 & 41.82 & 67.86 & 51.32 & 30.80 & 47.89 & 48.07 & 26.26 & 46.88 \\
ILCLIP & 36.68 & 12.96 & 32.78 & 33.71 & 8.02 & 22.36 & 41.24 & 6.05 & 16.65 \\
ConCLIP & 38.69 & 30.93 & 55.93 & 50.41 & 25.11 & 42.41 & 40.26 & 11.10 & 21.80 \\
NegCLIP & 60.95 & 47.81 & 73.21 & 49.90 & 36.92 & 56.12 & 55.19 & 31.22 & 56.99 \\
NegFullCLIP & 70.30 & 33.58 & 58.53 & 50.92 & 26.79 & 46.41 & 44.12 & 22.50 & 43.51 \\ 
OpenAI CLIP-B & 44.12 & 36.18 & 58.53 & 50.31 & 27.22 & 45.57 & 41.09 & 24.48 & 45.29 \\
OpenAI CLIP-L & 41.11 & 37.85 & 62.50 & 50.51 & 29.96 & 49.37 & 37.30 & 30.23 & 52.43 \\ \hline
LogicCLIP-Neg (Ours) & 85.73 & 51.21 & 76.27 & 60.39 & 33.33 & 57.81 & 64.97 & 33.20 & 56.99 \\
LogicCLIP-NegFull (Ours) & 85.63 & 50.00 & 75.29 & 60.29 & 33.97 & 56.54 & 67.70 & 31.42 & 56.59 \\
LogicCLIP-B (Ours) & 87.09 & 48.50 & 73.91 & 71.18 & 35.23 & 57.38 & 67.93 & 33.40 & 55.70 \\
LogicCLIP-L (Ours) & 79.25 & 52.71 & 78.86 & 74.54 & 37.34 & 56.54 & 67.78 & 35.98 & 59.56 \\
\bottomrule
\end{tabular}}
\caption{Performance comparison across different logical structures (Conjunction, Disjunction, and Negation).}
\label{tab:logical_comparison3}
\end{table*}

\begin{table*}[t]
\centering
\resizebox{0.9\textwidth}{!}{
\begin{tabular}{l|ccc|ccc|ccc}
\toprule
\textbf{Model} & \multicolumn{3}{c|}{\textbf{Contrast}} & \multicolumn{3}{c|}{\textbf{Comparison}} & \multicolumn{3}{c}{\textbf{Conditionality}} \\
\cmidrule(r){2-4} \cmidrule(r){5-7} \cmidrule(r){8-10}
& \textbf{MCQ} & \textbf{R@1} & \textbf{R@5} & \textbf{MCQ} & \textbf{R@1} & \textbf{R@5} & \textbf{MCQ} & \textbf{R@1} & \textbf{R@5} \\
\midrule
LaCLIP & 44.86 & 23.32 & 45.38 & 47.22 & 35.66 & 58.14 & 43.32 & 28.42 & 43.16 \\
LaionCLIP & 51.62 & 30.05 & 54.08 & 51.07 & 35.66 & 62.79 & 48.00 & 34.74 & 47.37 \\
DatacompCLIP & 39.73 & 7.63 & 19.50 & 35.97 & 8.53 & 24.03 & 33.96 & 7.37 & 15.79 \\
TripletCLIP-CC3M & 42.03 & 3.40 & 9.66 & 51.50 & 3.10 & 10.85 & 48.33 & 2.11 & 6.32 \\
TripletCLIP-CC12M & 59.64 & 9.18 & 21.88 & 56.10 & 13.18 & 25.58 & 51.22 & 7.37 & 14.74 \\
MetaCLIP & 52.94 & 28.26 & 51.58 & 49.57 & 30.23 & 56.59 & 45.10 & 23.16 & 47.37 \\
ILCLIP & 45.02 & 8.71 & 21.77 & 72.59 & 9.30 & 20.93 & 73.94 & 5.26 & 15.79 \\
ConCLIP & 39.68 & 22.30 & 42.81 & 41.86 & 24.81 & 50.39 & 39.42 & 21.05 & 45.26 \\
NegCLIP & 55.56 & 34.53 & 60.41 & 53.96 & 42.64 & 70.54 & 39.09 & 30.53 & 60.00 \\
NegFullCLIP & 56.92 & 23.91 & 45.74 & 60.39 & 29.46 & 51.94 & 45.77 & 26.32 & 42.11 \\ 
OpenAI CLIP-B & 36.32 & 24.75 & 44.36 & 40.36 & 28.68 & 51.94 & 32.74 & 21.05 & 42.11 \\
OpenAI CLIP-L & 37.42 & 32.02 & 52.47 & 36.83 & 32.56 & 56.59 & 26.61 & 20.00 & 49.47 \\ \hline
LogicCLIP-Neg (Ours) & 71.02 & 36.49 & 63.09 & 83.19 & 44.96 & 72.09 & 77.39 & 38.95 & 56.84 \\
LogicCLIP-NegFull (Ours) & 71.17 & 34.47 & 61.18 & 82.01 & 40.31 & 71.32 & 75.17 & 36.84 & 60.00 \\
LogicCLIP-B (Ours) & 74.48 & 34.35 & 60.47 & 90.69 & 41.86 & 68.22 & 86.75 & 33.68 & 63.16 \\
LogicCLIP-L (Ours) & 76.94 & 37.98 & 64.82 & 92.93 & 40.31 & 73.64 & 90.42 & 41.05 & 65.26 \\
\bottomrule
\end{tabular}}
\caption{Performance comparison across different logical structures (Contrast, Comparison, and Conditionality).}
\label{tab:logical_comparison4}
\end{table*}

\begin{table*}[t]
\centering
\resizebox{0.9\textwidth}{!}{
\begin{tabular}{l|ccc|ccc|ccc}
\toprule
\textbf{Model} & \multicolumn{3}{c|}{\textbf{Causality}} & \multicolumn{3}{c|}{\textbf{Temporality}} & \multicolumn{3}{c}{\textbf{Inclusion}} \\
\cmidrule(r){2-4} \cmidrule(r){5-7} \cmidrule(r){8-10}
& \textbf{MCQ} & \textbf{R@1} & \textbf{R@5} & \textbf{MCQ} & \textbf{R@1} & \textbf{R@5} & \textbf{MCQ} & \textbf{R@1} & \textbf{R@5} \\
\midrule
LaCLIP & 38.35 & 25.00 & 47.67 & 25.44 & 25.72 & 49.16 & 47.05 & 29.89 & 56.70 \\
LaionCLIP & 39.69 & 25.58 & 50.00 & 32.29 & 33.33 & 56.43 & 55.18 & 40.22 & 69.67 \\
DatacompCLIP & 34.23 & 8.14 & 15.70 & 25.88 & 8.01 & 21.75 & 39.53 & 9.45 & 27.25 \\
TripletCLIP-CC3M & 40.93 & 2.91 & 10.47 & 30.24 & 3.43 & 11.72 & 47.15 & 3.52 & 9.23 \\
TripletCLIP-CC12M & 48.35 & 8.72 & 21.51 & 32.29 & 9.29 & 23.70 & 57.22 & 10.33 & 30.33 \\
MetaCLIP & 40.00 & 25.58 & 46.51 & 40.70 & 31.25 & 54.34 & 56.00 & 38.24 & 66.37 \\
ILCLIP & 66.60 & 5.81 & 19.19 & 27.01 & 10.84 & 23.50 & 65.75 & 7.47 & 21.54 \\
ConCLIP & 31.03 & 22.67 & 37.79 & 23.94 & 22.83 & 45.39 & 40.14 & 21.10 & 42.64 \\
NegCLIP & 33.51 & 33.14 & 57.56 & 45.18 & 36.57 & 61.75 & 56.61 & 37.58 & 70.77 \\
NegFullCLIP & 40.93 & 23.26 & 44.19 & 47.22 & 24.85 & 46.40 & 68.60 & 28.35 & 56.26 \\ 
OpenAI CLIP-B & 20.10 & 24.42 & 42.44 & 19.46 & 24.18 & 45.32 & 50.51 & 30.11 & 58.90 \\
OpenAI CLIP-L & 19.59 & 27.91 & 49.42 & 20.38 & 34.41 & 54.95 & 45.02 & 36.04 & 64.62 \\ \hline
LogicCLIP-Neg (Ours) & 70.52 & 31.40 & 54.65 & 72.29 & 36.70 & 64.24 & 86.08 & 43.96 & 74.73 \\
LogicCLIP-NegFull (Ours) & 71.13 & 30.81 & 51.74 & 73.91 & 35.15 & 61.89 & 83.94 & 42.64 & 74.07 \\
LogicCLIP-B (Ours) & 79.79 & 27.91 & 48.26 & 81.46 & 36.43 & 61.01 & 91.87 & 45.27 & 74.07 \\
LogicCLIP-L (Ours) & 84.54 & 26.16 & 51.74 & 76.01 & 41.62 & 67.34 & 89.33 & 44.84 & 74.29 \\
\bottomrule
\end{tabular}}
\caption{Performance comparison across different logical structures (Causality, Temporality, and Inclusion).}
\label{tab:logical_comparison5}
\end{table*}

\subsection{C.6 Qualitative Analysis}
\label{SecC6}
\begin{figure*}[h]
\centering
	\includegraphics[trim=0 40 0 40, clip=true, width=0.75\textwidth]{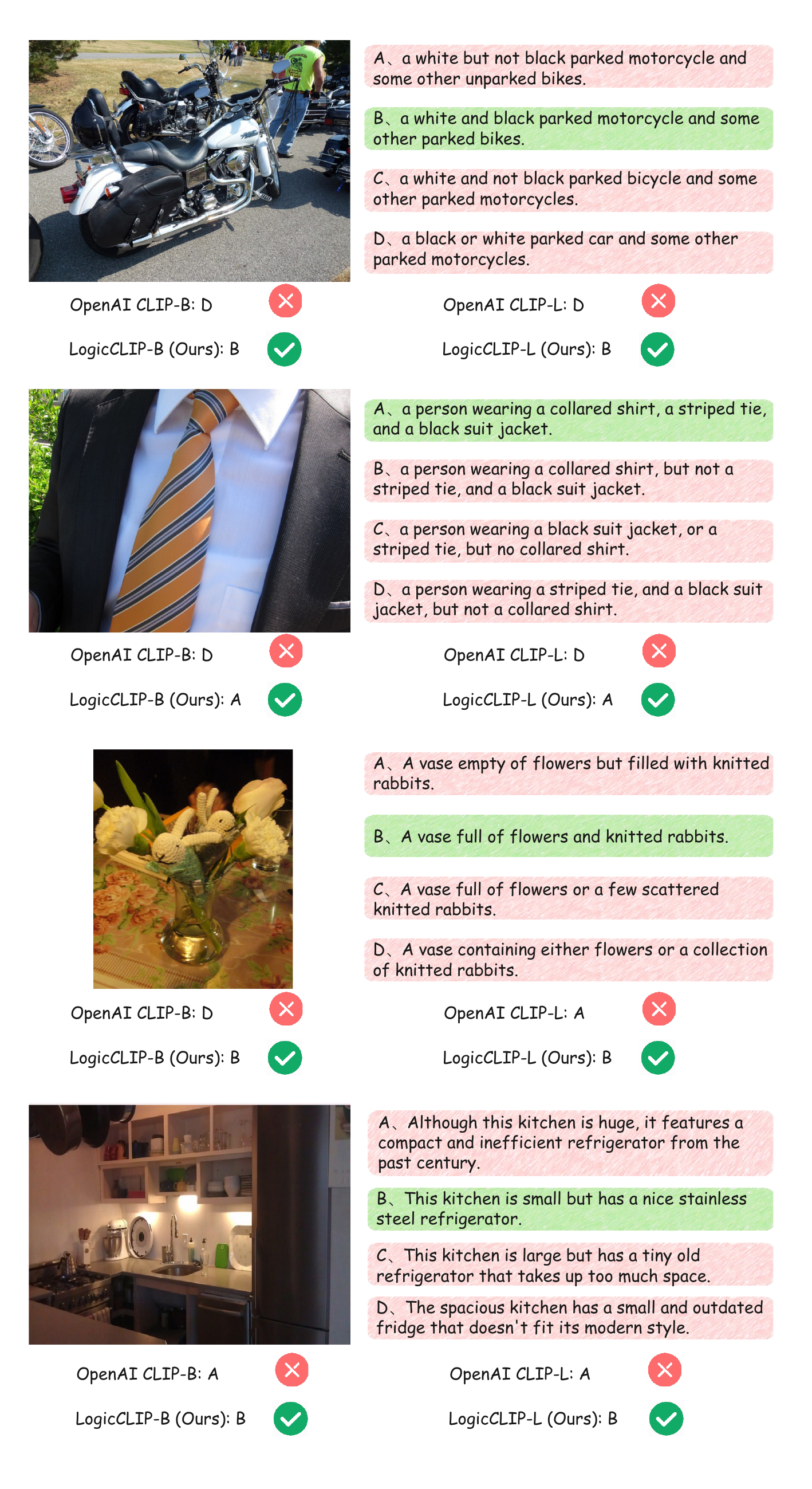}
	\caption{Visualization of LogicBench-Image Samples and Corresponding Model Performance.} 
 \label{fig6}
\end{figure*}

\begin{figure*}[h]
\centering
	\includegraphics[trim=0 40 0 40, clip=true, width=0.9\textwidth]{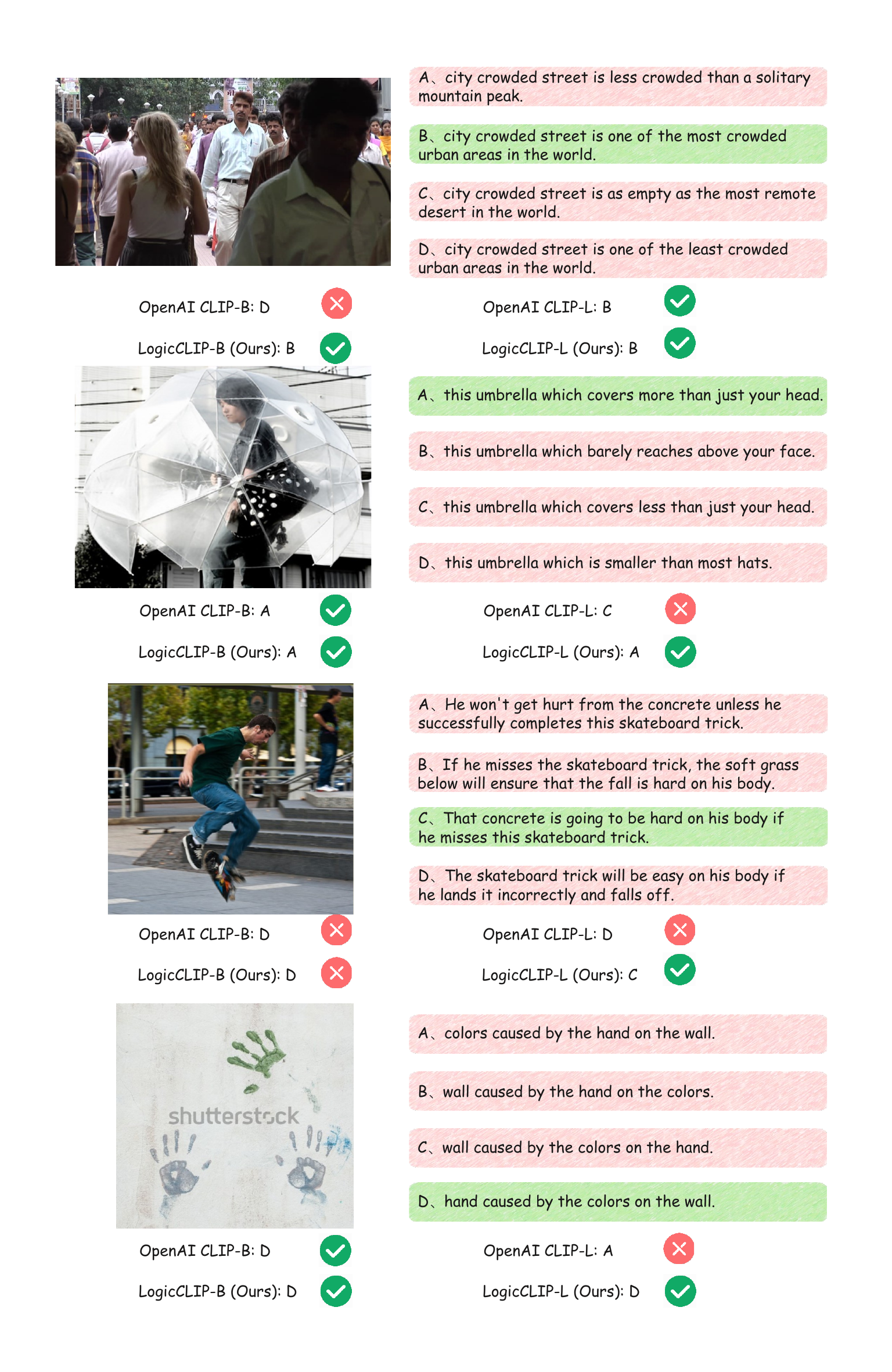}
	\caption{Visualization of LogicBench-Image Samples and Corresponding Model Performance.}
 \label{fig7}
\end{figure*}

\begin{figure*}[h]
\centering
	\includegraphics[trim=0 20 0 20, clip=true, width=0.9\textwidth]{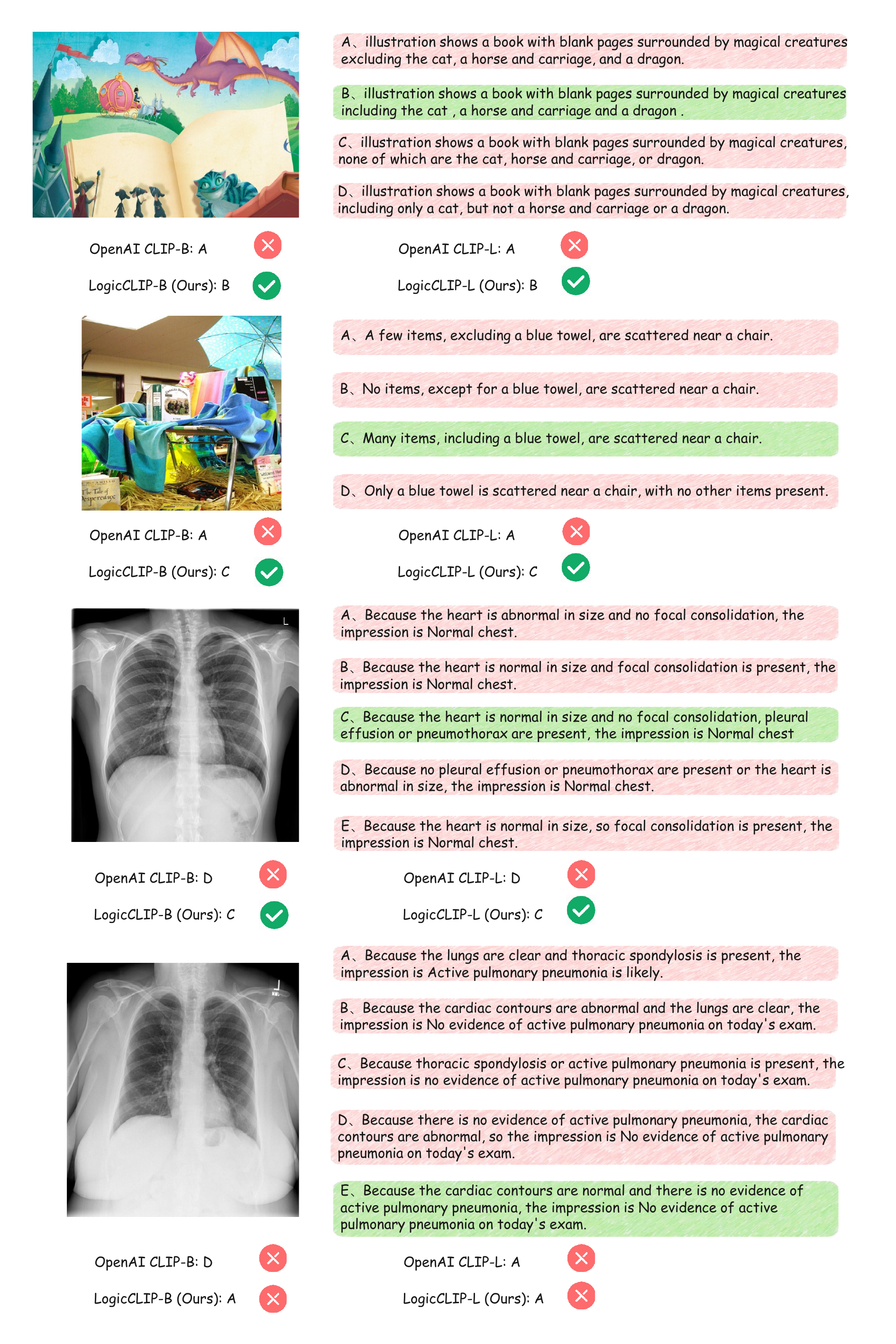}
	\caption{Visualization of Samples from LogicBench-Image and LogicBench-Medicine, and Corresponding Model Performance.}
 \label{fig8}
\end{figure*}

\begin{figure*}[h]
\centering
	\includegraphics[trim=40 120 40 80, clip=true, width=\textwidth]{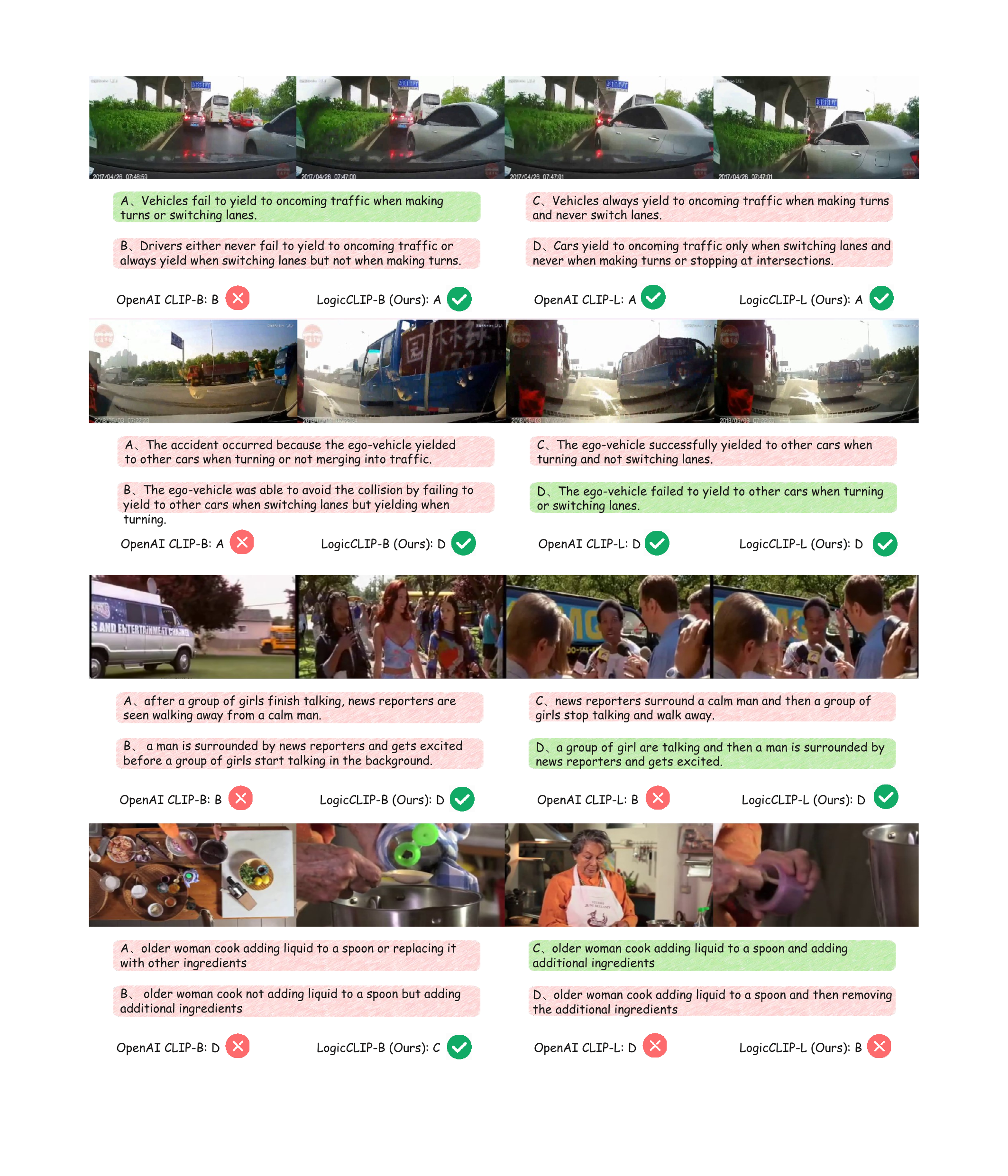}
	\caption{Visualization of Samples from LogicBench-Anomaly and LogicBench-Video, and Corresponding Model Performance.}
 \label{fig9}
\end{figure*}

To complement our quantitative findings, this section provides qualitative insights into LogicCLIP's enhanced logical understanding, contrasting its behavior with base CLIP models.
Figures \ref{fig6}, \ref{fig7}, \ref{fig8}, and \ref{fig9} present illustrative examples from LogicBench's image, video, anomaly, and medical scenarios, specifically focusing on the MCQ task.
These examples vividly demonstrate the ``logical blindspots" of base CLIP models. For instance, in Figure \ref{fig6} (top), base CLIP struggles to differentiate between ``including the cat, a horse and carriage, and a dragon" (correct) and logically perturbed negatives involving similar objects or negation, whereas LogicCLIP correctly identifies the nuanced logical structure. Similarly, Figure \ref{fig7} (middle) highlights LogicCLIP's ability to correctly choose ``this umbrella which covers more than just your head," demonstrating a grasp of comparative logic that eludes its base counterpart.

Furthermore, these visualizations also showcase LogicCLIP's robust generalization across diverse domains. As shown in Figure \ref{fig8}, two medical scenarios from chest X-ray reports are included. LogicCLIP correctly identifies the first causal statement, a task where the base CLIP model completely fails. However, since LogicCLIP was not specifically trained in the medical field, it still faces challenges with more complex, longer medical texts, even though its performance is superior to native CLIP.
Figure \ref{fig9} provides insights into Anomaly and Video scenarios. In the Anomaly example, LogicCLIP correctly discerns logical errors in the captions, while the base model selects a semantically similar but logically flawed option. 
The Video example further illustrates LogicCLIP's superior temporal and causal reasoning in a dynamic context. Collectively, these qualitative examples visually confirm that LogicCLIP’s training effectively enables it to move beyond superficial semantic matching, developing a deeper and more granular understanding of critical logical structures. This allows it to accurately differentiate between logically correct and incorrect descriptions even when semantic similarity is high, a capability absent in existing VLMs.

\end{document}